\documentclass[journal]{IEEEtran}
\usepackage{graphicx,amssymb,mathrsfs,amsmath,array,color}
\usepackage{subfigure}
\usepackage{multirow}
\usepackage{booktabs}
\usepackage{cite}
\usepackage[colorlinks,urlcolor=blue,driverfallback=dvipdfm]{hyperref}
\usepackage{makecell} 
\usepackage{pifont}
\newcommand{\cmark}{\ding{51}}
\newcommand{\xmark}{\ding{55}}
\usepackage{algorithm,algorithmicx}
\usepackage{algpseudocode}

\usepackage{nomencl}
\makenomenclature
\renewcommand\nomgroup[1]{\item[\bfseries]}

\begin{document}
\title{Multi-Source Collaborative Domain Generalization for Cross-Scene Remote Sensing Image Classification}

\author{Zhu Han,~\IEEEmembership{Student Member,~IEEE,}
        Ce Zhang,~\IEEEmembership{Member,~IEEE,}
        Lianru Gao,~\IEEEmembership{Senior Member,~IEEE,}
        Zhiqiang Zeng,~\IEEEmembership{Student Member,~IEEE,}
        Michael K. Ng,~\IEEEmembership{Senior Member,~IEEE,}
        Bing Zhang,~\IEEEmembership{Fellow,~IEEE,}
        and Jocelyn Chanussot,~\IEEEmembership{Fellow,~IEEE}

\thanks{This work was supported by the National Natural Science Foundation of China (NSFC) under Grant 42325104 and Grant 62161160336, and in part by the Joint NSFC-RGC under Grant N-HKU76921. (\emph{Corresponding author: Lianru Gao}).}
\thanks{Z. Han is with the Key Laboratory of Digital Earth Science, Aerospace Information Research Institute, Chinese Academy of Sciences, Beijing 100094, China, and with the International Research Center of Big Data for Sustainable Development Goals, Beijing 100094, China, and also with the College of Resources and Environment, University of Chinese Academy of Sciences, Beijing 100049, China (e-mail: hanzhu19@mails.ucas.ac.cn).}
\thanks{C. Zhang is with the School of Geographical Sciences, University of Bristol, Bristol BS8 1SS, UK, and with the UK Center of Ecology $\&$ Hydrology, Library Avenue, Lancaster LA1 4AP, UK (e-mail: ce.zhang@bristol.ac.uk).}
\thanks{L. Gao is with the Key Laboratory of Computational Optical Imaging Technology, Aerospace Information Research Institute, Chinese Academy of Sciences, Beijing 100094, China (e-mail: gaolr@aircas.ac.cn).}
\thanks{Z. Zeng is with the Beijing Institute of Remote Sensing Equipment, Beijing 100854, China (e-mail: zengzq@buaa.edu.cn).}
\thanks{Michael K. Ng is with the Department of Mathematics, Hong Kong Baptist University, Hong Kong, China (e-mail: michael-ng@hkbu.edu.hk).}
\thanks{B. Zhang is with the Aerospace Information Research Institute, Chinese Academy of Sciences, Beijing 100094, China, and also with the College of Resources and Environment, University of Chinese Academy of Sciences, Beijing 100049, China (e-mail: zhangbing@aircas.ac.cn).}
\thanks{J. Chanussot is with the Univ. Grenoble Alpes, CNRS, Grenoble INP, IJK, Grenoble 38000, France, and also with the Aerospace Information Research Institute, Chinese Academy of Sciences, 100094 Beijing, China (e-mail: jocelyn.chanussot@inria.fr).}
}

\markboth{Accepted by IEEE Transactions on Geoscience and Remote Sensing,~Vol.~XX, No.~XX, ~XXXX,~2024}
{Han \MakeLowercase{\textit{et al.}}: }
\maketitle

\begin{abstract}
Cross-scene image classification aims to transfer prior knowledge of ground materials to annotate regions with different distributions and reduce hand-crafted cost in the field of remote sensing. However, existing approaches focus on single-source domain generalization to unseen target domains, and are easily confused by large real-world domain shifts due to the limited training information and insufficient diversity modeling capacity. To address this gap, we propose a novel multi-source collaborative domain generalization framework (MS-CDG) based on homogeneity and heterogeneity characteristics of multi-source remote sensing data, which considers data-aware adversarial augmentation and model-aware multi-level diversification simultaneously to enhance cross-scene generalization performance. The data-aware adversarial augmentation adopts an adversary neural network with semantic guide to generate MS samples by adaptively learning realistic channel and distribution changes across domains. In views of cross-domain and intra-domain modeling, the model-aware diversification transforms the shared spatial-channel features of MS data into the class-wise prototype and kernel mixture module, to address domain discrepancies and cluster different classes effectively. Finally, the joint classification of original and augmented MS samples is employed by introducing a distribution consistency alignment to increase model diversity and ensure better domain-invariant representation learning. Extensive experiments on three public MS remote sensing datasets demonstrate the superior performance of the proposed method when benchmarked with the state-of-the-art methods.

\end{abstract}
\graphicspath{{figures/}}


\nomenclature[1]{$\mathcal{X}$, $\mathcal{\hat{X}}$}{Set of input and augmented MS data.}
\nomenclature[2]{$\mathbf{X}$, $\mathbf{\hat{X}}$}{Input and augmented SD data.}
\nomenclature[3]{$\mathbf{p}^{g}$}{Predicted probability output by adversarial augmentation.}
\nomenclature[4]{$\mathbf{y}$}{Labels of MS data.}
\nomenclature[5]{$N_{S}$}{Number of SD samples.}
\nomenclature[6]{$\mathbf{F}^{spa}$, $\mathbf{F}^{cha}$}{Spatial and channel features extracted by domain encoder.}
\nomenclature[7]{$\mathbf{F}_{cross}$, $\mathbf{F}_{intra}$}{Cross-domain and intra-domain features by fusing information between SDs.}
\nomenclature[8]{$\mathbf{P}$}{Cross-domain class-wise prototypes.}
\nomenclature[9]{$\phi_{c}$}{Intra-domain class modeling by KMM.}
\printnomenclature[3cm] 

\section{Introduction}
\IEEEPARstart{W}{ith} the significant improvement in imaging technology on space-borne and airborne platforms, a large amount of remote sensing images are gradually available and provide rich practical values in Earth observation. Equipped by the excellent capability in automatically extracting inherent ground information and discriminating various categories, deep learning-based remote sensing image classification has become increasingly crucial for its broad real-world applications, such as land use\cite{ma2017review, zhang2019joint}, agriculture monitoring\cite{ ratnayake2021towards, han2023spatio} and urban planing\cite{srivastava2019understanding, qiu2019local}. However, most state-of-the-art image classification approaches are based on supervised training paradigms and heavily rely on numerous high-quality hand-crafted labeled data. It is unrealistic to obtain sufficient manually-labeled data for all circumstances and the direct migration of well-trained models also presents the challenge of degraded generalization performance due to domain shift. To alleviate the cumbersome data labeling requirement, cross-scene image classification technologies have been explored by transferring prior knowledge of ground objects learned from existing remote sensing datasets to annotate unseen scenarios\cite{othman2017domain}, further providing potential for recognizing ground categories of new scenes in actual applications.

As the majority representatives of cross-scene image classification, domain adaptation (DA) remedies the issue of domain shift by learning domain-invariant features between well-labeled source domain (SD) and unlabeled target domain (TD) \cite{chen2020domain, wu2021heterogeneous, peng2022domain}. Owing to the availability of target samples, the mainstream solutions aim to minimize distribution discrepancy of different domains within the complicated features or output spaces by setting various metrics\cite{Tsai_2019_ICCV, sharma2021instance}, such as maximum mean discrepancy (MMD)\cite{long2019deep, zhu2020deep}, optimal transport distance\cite{flamary2016optimal, zhang2021topological} and adversarial learning based distribution alignment \cite{8099799, yu2019transfer, liu2020class}. Despite the remarkable performance and efficiency demonstrated by these DA-based methods, DA relies on the strong assumption that target data is accessible for model adaptation and it is hard to acquire target data from real-world scenarios. This issue inspires the research area of domain generalization (DG)\cite{zhou2022domain}, whose goal is to learn the model trained on single or multiple seen domains to achieve accurate predictions on unseen TD. It is especially difficult because of the limited information available to train the model with SD data only. Therefore, most existing DG methods utilize adversarial training\cite{li2018deep, li2018domain}, self-supervised\cite{carlucci2019domain, kim2021selfreg}, meta learning \cite{balaji2018metareg, li2018learning, 10287966} or domain augmentation \cite{shankar2018generalizing, zhou2020learning, li2021progressive} techniques to better learn domain-invariant representations from the perspective of expanding valuable samples or modeling agnostic training, and have shown promising classification performance. 

\begin{figure}[!t]
	  \centering
		\subfigure{
			\includegraphics[width=0.47\textwidth]{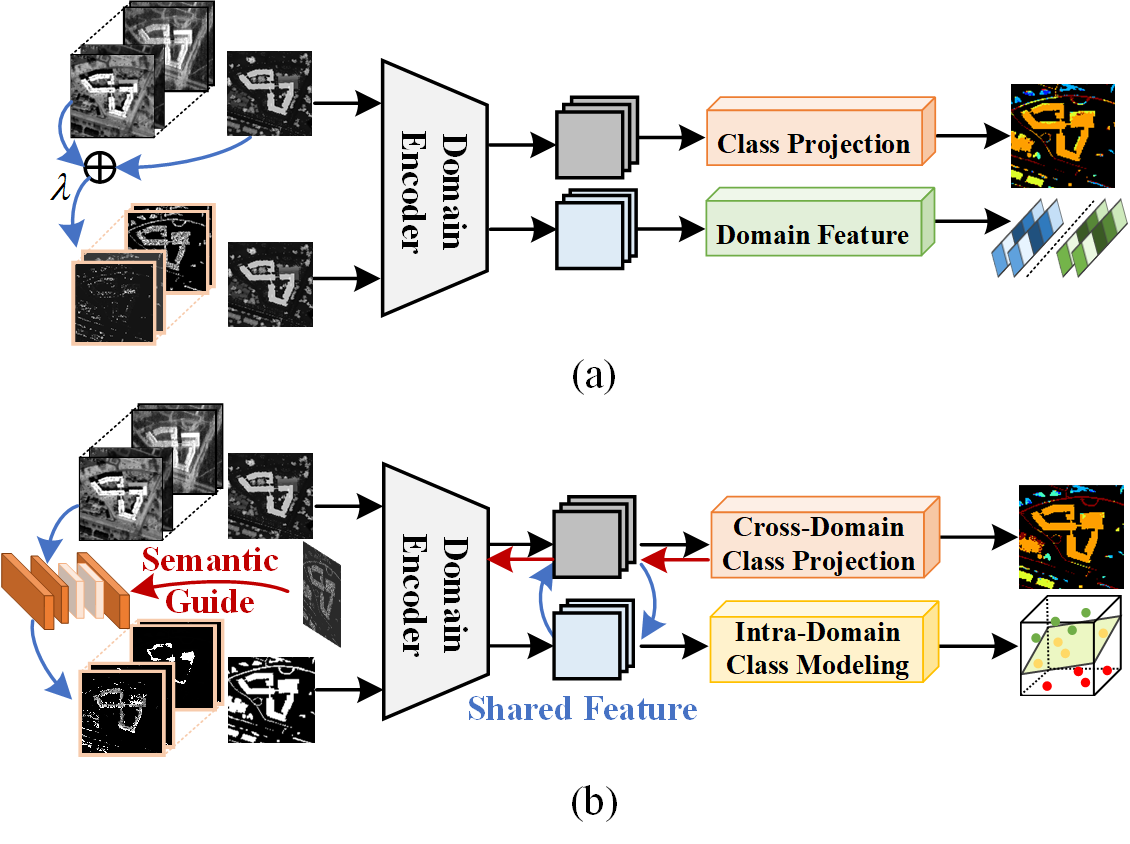}
		}
        \caption{Illustrative comparison for MS data augmentation and diversity modeling on different cross-scene methods. (a) Existing cross-scene framework. (b) Our proposed MS-CDG framework.}
\label{fig:motivations}
\end{figure}

Compared with natural images, there are more challenging cross-domain influences in the field of remote sensing, including sensor, angle, light, weather and season \cite{sinha2012seasonal, teng2019classifier}, which can be summarized as domain style differences. It refers to the inherent characteristics of a particular domain covering distribution information in the feature and label space, which are often influenced by imaging conditions and prior shifts across different sensors or geographic areas \cite{MA2024113924}. Therefore, the direct application of DG methods in remote sensing scenarios will eliminate a part of discriminative features in the feature regularization process and may over-learn the statistical distribution of the given SD owing to ignoring inherent ground characteristics, especially for ground objects with different styles or imbalanced sample proportions \cite{zhu2023style, han2024dual}. Some pioneers tackle this problem by encoding the semantic and morphological structure of ground objects in the generated domain expansion and significantly enhance the extraction performance of domain-invariant features compared with traditional DA-based methods \cite{zhang2023single, 10268956}. However, these approaches usually concentrate on single-source DG and the obtained domain-invariant features are incomplete without considering essential class semantic information, which cannot solve this dilemma in regarding to large real-world domain shifts between remote sensing observations over distant regions. In this context, the emergency of a variety of image products, e.g., hyperspectral image (HSI), multispectral image, synthetic aperture radar (SAR), light detection and ranging (LiDAR) sensors, can provide rich spectral, spatial or channel information about various geographic objects or materials in the scene that is unachievable by single red-green-blue (RGB) image \cite{hong2020more, han2022multimodal}. The existence of multi-source (MS) data also brings huge challenges for dealing with multi-dimensional modality blending problem while maintaining diverse class information, which limits the efficiency and stability of cross-scene image classification technologies.

To promote the development of precise cross-scene image classification in the remote sensing era, narrowing the distribution disparity and class difference between different SDs becomes a necessity by the combination of MS collaborate characteristics and high-performance diversity modeling. However, as illustrated in Fig. \ref{fig:motivations}, the generalization of ordinary DG models to cross-scene image classification task is limited and has the following drawbacks. First, mainstream DG-based cross-scene classification approaches focus on learning projected features and class prediction independently through disentanglement, which makes the model overemphasize the semantic invariance across domain and weaken the class representation inevitably. More precisely, they tend to prioritize variations caused by domain biases, while the inherent cross-domain class invariance and class-specific modeling within each domain are easily neglected. As a result, some hard ground objects with similar domain styles could be misclassified due to the lack of class guidance information. Second, from the perspective of data augmentation, existing domain expansion strategies merely adopt pre-defined out-of-domain styles to augment SD samples based on the data input level, such as mixup \cite{zhang2017mixup} or particular types of corruption augmentations \cite{geirhos2018generalisation}, and fail to capture sufficient diversity induced by the simultaneous appearance of changes in both the channel and distribution space. MS remote sensing data generally have certain texture homogeneity and channel heterogeneity characteristics related to the imaging mechanism, which is essential for extracting complete domain-invariant representations. Nevertheless, current data augmentation techniques do not take into account the uniqueness of imaging attributes within MS data, further widening the domain gap for subsequent model training and leading to performance degradation. Therefore, constructing suitable multi-domain attribute guidance for data augmentation can help generate more reliable MS images to enhance generalization performance.

To this end, based on homogeneity and heterogeneity characteristics of MS remote sensing data, a novel multi-source collaborative domain generalization (MS-CDG) framework is proposed to mitigate the aforementioned issues in the cross-scene image classification task. We firstly revisit the cross-scene problem from the perspective of MS and DG in the remote sensing community, targeting MS data acquired by different imaging sensors. Specifically, MS-CDG considers two different views to enhance cross-scene generalization performance: data-aware adversarial augmentation and model-aware multi-level diversification. The data-aware adversarial augmentation applies an adversary neural network to adaptively learn both band-by-band semantic changes and domain information, and finally generate the semantic-aware multi-domain augmented samples to improve the generalization ability. Based on the spatial and channel dimension of MS data, the model-aware multi-level diversification simultaneously considers the long-distance cross-domain and short-distance intra-domain modeling mechanism by the class-wise prototype and kernel mixture module respectively, which enables MS-CDG to capture the most discriminative and higher-order class information over various SDs, while maintaining a moderate complexity with respect to multi-dimensional object characteristics. Furthermore, the joint classification of original and novel derived MS samples is explored by the distribution consistency alignment to increase model diversity and ensure better domain-invariant representation learning. In brief, the major contributions of this paper can be summarized as follows.
\begin{itemize}
    \item A novel DG framework based on MS data is proposed for cross-scene classification, considering both the data-aware adversarial augmentation and model-aware multi-level diversification to enhance robust and generalization performance on complex MS remote sensing scenarios.
    \item The data-aware adversarial augmentation achieves multi-domain augmentation on cross-domain images, which maintains certain semantic consistency and domain styles by performing adversarial training in the function space of a partly weight-sharing adversary neural network.
    \item To explore the fine-grained DG for each modality, the model-aware multi-level diversification is devised, which helps to aggregate more multi-dimensional object characteristics and make the class modeling more complete.
    \item The distribution consistency constraint is developed to ensure high-quality training across original and augmented samples, which further achieves stable joint classification and improves the generalization capability for unseen domains.
\end{itemize}



\begin{figure*}[!t]
	  \centering
		\subfigure{
			\includegraphics[width=0.95\textwidth]{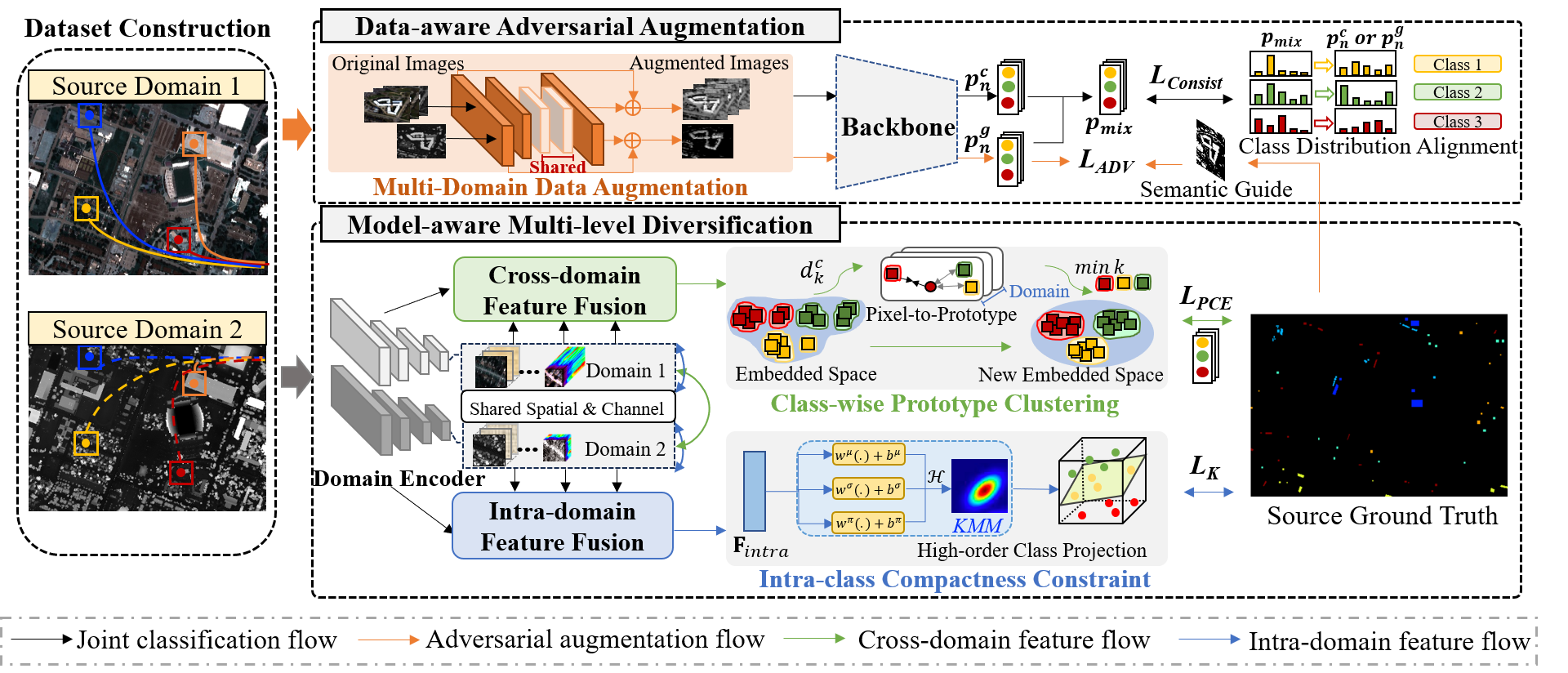}
		}
        \caption{The framework of the proposed MS-CDG, including data-aware adversarial augmentation and model-aware multi-level diversification. The multi-domain data augmentation is designed by a partly weight-sharing adversary neural network with the semantic guide to generate the mixed MS images, and further fed into the trained backbone to achieve joint classification. The embedding features of cross-domain and intra-domain levels are simultaneously optimized by modeling pixel-to-prototype and high-order intra-class compactness relationships for different domains to enhance domain-invariant representation capability.}
\label{fig:workflow}
\end{figure*}

The remaining part of this paper is organized as follows. Section II introduces the important works related to our MS-CDG. In Section III, we elaborate the implementation details of our MS-CDG. Section IV presents the extensive experiments and analyses on three MS remote sensing datasets. Finally, conclusions are drawn in Section V.

\section{Related Work}

\subsection{Multi-Source Domain Generalization (MSDG)}
Different from single-source DG, MSDG aims to capture the knowledge of multiple SDs for training and make accurate prediction on other unseen TDs. MSDG usually assumes different distinct but relevant SDs are available. Along this line, current approaches mainly leverage various clues inside multiple domains to design suitable MSDG models, such as domain-invariant feature learning \cite{ding2018, NEURIPS2021_a8f12d94},  feature fusion \cite{shen2019situational}, self-supervised \cite{carlucci2019domain, wang2020learning} and meta-learning paradigm \cite{li2018learning, dou2019domain}. For instance, Ding et al. \cite{ding2018} developed an effective deep structure with the structured low-rank constraint to effectively separate domain-specific and domain-invariant components across multiple sources. Wang et al. \cite{wang2020learning} leveraged the self-supervised training paradigm Jigsaw Puzzle and memory bank to jointly learn the spatial relationships of images and enhance the generalization capacity of the model. To capture generalizable features, Dou et al. \cite{dou2019domain} proposed a meta-learning strategy to model the meta-features for DG and introduced semantic guidance of feature space to prevent domain shift, but the complex meta-learning procedure makes optimization difficult. 

Considering this drawback, another prominent direction of MSDG is to enrich the domain diversity of training data by introducing various forms of data augmentation. More precisely, Shankar et al. \cite{shankar2018generalizing} utilized the concept of adversarial attack and defence, and proposed a cross-gradient method (CrossGrad) by sampling domain-guided perturbation to augment the training data. Zhou et al. \cite{zhou2020learning} designed a conditional generator network by learning the distribution divergence between source and synthesized pseudo-novel domains to generate images with significantly different distributions. As the typical representatives of data augmentation, mixup and its variants \cite{xu2020adversarial, zhang2017mixup, shu2021open, kim2021selfreg} can effectively deal with domain-level and class-level interpolated samples by performing empirical risk minimization to augment SDs. However, most existing methods only consider the pre-defined mixup ratio distribution to interpolate augmented samples, e.g., beta distribution, and they fail to consider the multi-domain level mixup problems under different channel settings. Unlike natural images, it is hard to completely assess the semantics and diversity of MS remote sensing data owing to its texture homogeneity and channel heterogeneity characteristics from different imaging sensors. Past works on MSDG have not investigated both the channel and distribution changes for data augmentation across domains, especially for cross-scene image classification application. This paper has shown that the multi-domain data augmentation plays a crucial role in MSDG and it can improve the generalization performance on standard benchmarks.

\subsection{Cross-Scene Image Classification in Remote Sensing}
Cross-scene image classification is a specific cross-domain classification task that focuses on transferring limited training samples to learn valuable ground attributes and reduce label annotation on unlabeled scenarios, owing to the differences of imaging environments and geographical areas in the field of remote sensing. DA and DG are two main techniques to deal with the cross-scene image classification task. DA-based methods usually assume that the SD and TD have the same feature and label space but exist a certain marginal or conditional distribution differences. Therefore, benefited from the introduction of TD samples, current DA-based approaches mainly focus on learning the complex distribution variation between domains through feature or distribution alignment strategies, such as deep adaptation network (DAN) \cite{long2019deep}, deep subdomain adaptation network (DSAN) \cite{zhu2020deep} and topological structure and semantic information transfer network (TSTNet) \cite{zhang2021topological}. In addition, some generative and adversarial DA strategies are employed to deal with the domain shift problem and realize domain feature alignment, including adversarial discriminative domain adaptation (ADDA) \cite{8099799}, dynamic adversarial adaptation network (DAAN) \cite{yu2019transfer} and class-wise distribution adaptation (CDA) \cite{liu2020class}. Huang et al.\cite{9924236} proposed a novel two-branch attention adversarial domain adaptation (TAADA) network for HSI classification, which embedded DA into the process of adversarial learning to learn discriminative and domain-invariant features and effectively solved the distribution discrepancy problem. MDA-Net \cite{zhang2023cross} further introduced the collaborate MS information to boost cross-scene classification performance. However, these DA-based methods rely on SD and TD prior knowledge during adaptation, which further limits the generalization performance when the labeled target data are unavailable in practical scenarios.

The DG-based methods are more challenge than DA because of the limited training information and complex ground attributes of remote sensing images. SDENet \cite{zhang2023single} firstly considered the single-source DG problem in the realm  of cross-scene image classification, and effectively mined both semantic and morph information of HSI to extract domain-invariant representations. Zhao et al. \cite{10268956} proposed a locally linear unbiased randomization network (LLURNet) to achieve domain expansion with local style randomization to avoid duplicate augmentation and enhance generalization performance. The introduction of the language-mode prior knowledge can further provide helpful guidance for existing cross-scene models \cite{zhang2023language}. Nevertheless, with the help of contrastive learning paradigm, mainstream DG-based approaches focus on independent feature regularization of projected style features and classification prediction, which inevitably weakens the representation capability of the model in the disentanglement process. This deterministic prediction mode overemphasizes the semantic invariance across domains, and generates hard-coded features, so that the training process of the model lacks flexibility in representing class dependencies. In this paper, we design a model-aware multi-level diversification module by transforming embedded features into class-wise prototypes and a mixture of exponential kernels, respectively, to account for domain discrepancies and class dependencies in a different way.



\section{Methodology}
\subsection{Overview of MS-CDG}
As illustrated in Fig. \ref{fig:workflow}, our MS-CDG consists of two parts: data-aware adversarial augmentation, and model-aware multi-level diversification considering both cross-domain and intra-domain class modeling. The data-aware adversarial augmentation aims at generating the augmented MS images by training a partly weight-sharing adversary neural network to perform multi-domain data augmentation of inputs from different channels and further achieve joint classification. Based on the shared spatial and channel information of MS images extracted by domain encoder, the model-aware multi-level diversification aggregates multi-dimensional domain-invariant characteristics by modeling pixel-to-prototype and high-order intra-class relationships for different domains to improve the generalization capability. 

\subsection{Data-aware Adversarial Augmentation}
Given the input MS images $\mathcal{X}=\left\{\mathbf{X}_{k}\right\}_{k=1}^{2}\in \mathbb{R}^{H\times W\times N_{k}}$, containing $HW$ pixels in the spatial domain and $N_{k}$ channels in the $k$-th SD, a partly weight-sharing adversary neural network is designed in MS-CDG to generate the augmented MS images $\mathcal{\hat{X}}=\left\{\mathbf{\hat{X}}_{k}\right\}_{k=1}^{2}\in \mathbb{R}^{H\times W\times N_{k}}$, which can be formulated as:
\begin{equation}
\label{eq1}
\begin{aligned}
    \mathbf{h}_{k}^{(e)} = \left\{
    \begin{array}{ll}
    g(T_{k}^{(e)}(\mathbf{X}_{k})), & e = 1 \\
    g(T_{k}^{(e)}(\mathbf{h}_{k}^{(e-1)})), & e = 2 \\
    g(T_{shared}(\mathbf{h}_{k}^{(e-1)})), & e = 3,4 \\
    \mathbf{X}_{k}+T_{k}^{(e)}(\mathbf{h}_{k}^{(e-1)}), & e = 5 \\
    \end{array}
    \right.
\end{aligned}
\end{equation}
where $\mathbf{h}_{k}^{(e)}$ is the extracted hierarchical representation of adversary neural network in the $e$-th layer. $T_{k}(\cdot)$ and $T_{shared}(\cdot)$ represent the independent and shared $3\times 3$ convolution operation for different SDs, respectively. $g(\cdot)$ denotes the LeakyReLU activation function. The clamp function is applied to the output of the last layer $\mathbf{h}_{k}^{(5)}$ to enforce all elements into the range [0,1] and further generate the augmented MS images $\mathbf{\hat{X}}_{k}$. The architecture of adversary neural network in Eq.(\ref{eq1}) adopts a data-driven strategy to generate samples by introducing shared layers, so that shared domain features can be learned in the training process.

Since the original MS remote sensing data has different channel dimensions and ground attribute distributions, the direct training of the adversary neural network without relying on any prior knowledge fails to capture homogeneity and heterogeneity properties within MS data. Therefore, the adversarial loss $L_{ADV}$ is designed by imposing a learned semantic guide, which is achieved by maximizing the cross-entropy (CE) loss between the predicted probability output $\mathbf{p}_{n}^{g}\in \mathbb{R}^{C}$  and one-hot label $\mathbf{y}_{n}\in \mathbb{R}^{C}$, and minimizing the total variation (TV) loss on the generated MS images.
\begin{equation}
\label{eq2}
\begin{aligned}
    L_{ADV} &= -L_{CE}(\mathbf{p}_{n}^{g},\mathbf{y})+L_{TV}(\mathbf{\hat{X}}_{1})+L_{TV}(\mathbf{\hat{X}}_{2}) \\
    &=\frac{1}{N_{S}}\sum_{n=1}^{N_{S}}\mathbf{y}_{n}\log(\mathbf{p}_{n}^{g}) +\sum_{k=1}^{2}\sum_{i=1}^{H}\sum_{j=1}^{W}((\mathbf{\hat{x}}_{i,j+1}^{k}-\mathbf{\hat{x}}_{i,j}^{k})^{2} \\
    &+ (\mathbf{\hat{x}}_{i+1,j}^{k}-\mathbf{\hat{x}}_{i,j}^{k})^{2})
\end{aligned}
\end{equation}
where $N_{S}$ represents the number of SD samples. $\mathbf{\hat{x}}_{i,j}^{k}\in \mathbb{R}^{N_{k}}$ denotes the pixel of the generated image in the $k$-th SD. The optimization of $L_{ADV}$ can help the adversary neural network generate realistic image transformations with more semantic changes and better suppress the noise of MS images. The partly weight-sharing architecture of adversary neural network can further facilitate the heterogeneous feature learning of channel dimension during the optimization process.

\begin{figure}[!t]
	  \centering
		\subfigure{
			\includegraphics[width=0.45\textwidth]{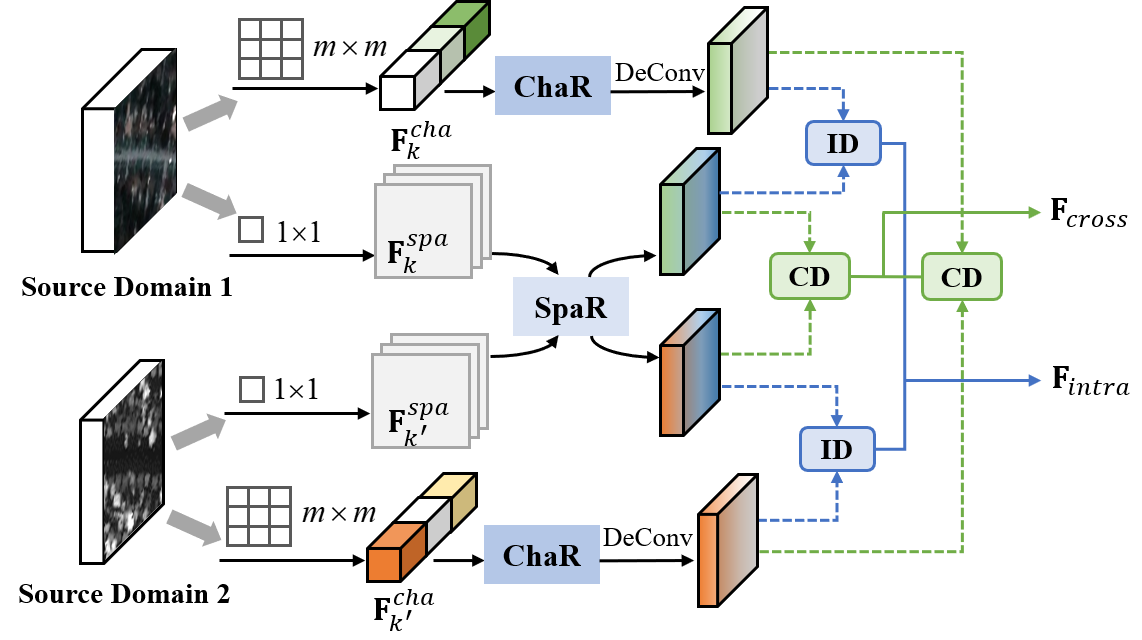}
		}
        \caption{Architecture of the domain encoder consisting of spatial randomization (SpaR) and channel randomization (ChaR) for different domains. The generated multi-dimensional domain information is further sent to achieve intra-domain (ID) and cross-domain (CD) feature fusion.}
\label{fig: domain module}
\end{figure}

\subsection{Model-aware Multi-level Diversification}
For the MS remote sensing images, the homogeneity and heterogeneity characteristics are vital for extracting complete domain-invariant representation. However, there is a certain correlation between the domain information and class of ground objects in different SDs, which inevitably leads to the possibility that different classes have similar diagnostic characteristics and domain discrepancies. Accordingly, reducing the negative impact of uncertainty between class and domain is crucial for the training of domain-invariant features. The motivation of MS-CDG is to effectively aggregate the domain and class correlations of existing training data through the parallel design of cross-domain and intra-domain class modeling, so that the model can be regularized to facilitate knowledge transfer and enhance cross-scene generalization performance.

\subsubsection{Domain Encoder}
Considering the cubic structure of MS data, the spatial and channel dimension information is encoded to capture differences in geometry and imaging properties. The training of MS-CDG adopts the $m\times m\times N_{k}$ image patch as the basic unit for different SDs. Firstly, the encoded spatial features $\mathbf{F}^{spa}_{k}\in \mathbb{R}^{m\times m\times d_{spa}}$ and channel features $\mathbf{F}^{cha}_{k}\in \mathbb{R}^{1\times 1\times d_{cha}}$ are obtained using $1\times 1$ and $m\times m$ convolution kernels respectively, as illustrated in Fig. \ref{fig: domain module}. Compared with the spatial dimension, the information that reflects the imaging attributes in the channel dimension is significantly distinct. Therefore, in order to improve the robustness of the imaging style in the encoding process, Adaptive Instance Normalization (AdaIN) \cite{karras2019style} for multi-dimensional features is designed to implement randomization to mitigate the style bias in different SDs. The spatial randomization (SpaR) aims to interact different SD features to learn domain-generalized semantic information and reduce imaging attribute differences. The channel randomization (ChaR) introduces certain channel content to enrich channel features by randomly selecting $\mathbf{F}^{'cha}_{k}$ from minibatch. The encoding processing of SpaR and  ChaR can be formulated as follows:
\begin{equation}
\label{eq3}
    {\rm SpaR}(\mathbf{F}^{spa}_{k}) = \sigma(\mathbf{F}^{spa}_{k'})\frac{\mathbf{F}^{spa}_{k}-\mu(\mathbf{F}^{spa}_{k})}{\sigma(\mathbf{F}^{spa}_{k})+\epsilon} + \mu(\mathbf{F}^{spa}_{k'})
\end{equation}
\begin{equation}
\label{eq4}
    {\rm ChaR}(\mathbf{F}^{cha}_{k}) = \sigma(\mathbf{F}^{cha}_{k})\frac{\mathbf{F}^{'cha}_{k}-\mu(\mathbf{F}^{'cha}_{k})}{\sigma(\mathbf{F}^{'cha}_{k})+\epsilon} + \mu(\mathbf{F}^{cha}_{k})
\end{equation}
where $\mu(\mathbf{F}^{spa}_{k})\in \mathbb{R}^{d_{spa}}$ and $\sigma(\mathbf{F}^{spa}_{k})\in \mathbb{R}^{d_{spa}}$ are the spatial-wise mean and standard deviation. $\mu(\mathbf{F}^{cha}_{k})\in \mathbb{R}^{d_{cha}}$ and $\sigma(\mathbf{F}^{cha}_{k})\in \mathbb{R}^{d_{cha}}$ are the channel-wise mean and standard deviation. $\epsilon$ is a small value to avoid division by zeros. After obtaining $\mathbf{F}_{k}^{cha}$, the deconvolution operation is adopted to restore the image patch size.

\begin{figure}[!t]
	  \centering
		\subfigure{
			\includegraphics[width=0.52\textwidth]{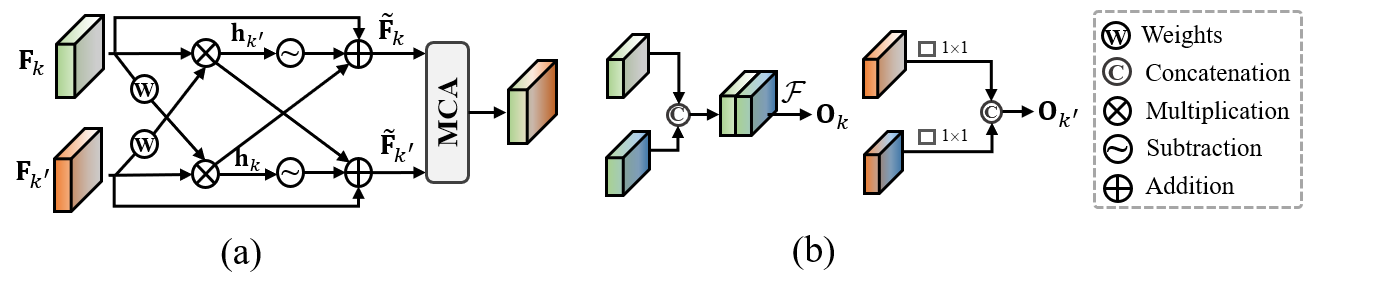}
		}
        \caption{The conceptual illustration of different feature fusion strategies. (a) Cross-domain feature fusion. (b) Intra-domain feature fusion.}
\label{fig: CDID}
\end{figure}

\subsubsection{Cross-domain Class-wise Prototype Clustering}
In view of advantageous long-distance modeling capability, the multi-head cross-attention (MCA) structure with the coupled enhancement module is employed to refine the multi-dimensional domain information and achieve cross-domain feature fusion, as shown in Fig. \ref{fig: CDID} (a). The coupled enhancement module adopts a residual learning mechanism to enhance the interaction of $k$-th and $k'$-th SD by adaptively generating complementary components $\mathbf{h}_{k}$ and $\mathbf{h}_{k'}$, which is written as:
\begin{equation}
\label{eq5}
    \mathbf{h}_{k'}=\mathbf{F}_{k}\cdot \lambda_{k}\cdot {\rm sigmoid}(\mathcal{W}(\mathbf{F}_{k'}))
\end{equation}
\begin{equation}
\label{eq6}
    \mathbf{\tilde{F}}_{k}=\mathbf{F}_{k}+\mathbf{h}_{k}-\mathbf{h}_{k'}
\end{equation}
where $\mathcal{W}$ refers to the point-wise convolution with $1\times 1$ kernel. $\lambda_{k}$ is a learnable coefficient to regulate information propagation for the spatial and channel dimension. The sigmoid function is adopted to generate coefficients between 0 and 1. Given $\mathbf{\tilde{F}}_{k}$ and $\mathbf{\tilde{F}}_{k'}$, the output of MCA with the residual shortcut is defined as follows.
\begin{equation}
\label{eq7}
{\rm MCA}(\mathbf{\tilde{F}}_{k},\mathbf{\tilde{F}}_{k'})={\rm Concat}(head_{i},\cdots,head_{h})\mathbf{W}+\mathbf{\tilde{F}}_{k}
\end{equation}
\begin{equation}
\label{eq8}   head_{i}={\rm softmax}(\frac{\mathbf{Q}_{c}\mathbf{K}_{c}^{T}}{\sqrt{D/h}})\mathbf{V}_{c}
\end{equation}
where $\mathbf{W}\in \mathbb{R}^{hd_{v}\times d_{k}}$ is the linear transformation matrice used for feature space transformation. $\mathbf{Q}_{c}=\mathbf{\tilde{F}}_{k}\mathbf{W}_{c}^{Q}$, $\mathbf{K}_{c}=\mathbf{\tilde{F}}_{k'}\mathbf{W}_{c}^{K}$, and $\mathbf{V}_{c}=\mathbf{\tilde{F}}_{k'}\mathbf{W}_{c}^{V}$ represent the query, key and value for the MCA mechanism, respectively. $D$ and $h$ are the embedding dimension and number of heads.

We concatenate the output together in the spatial and channel dimension, and then aggregate them using $3\times 3$ and $1\times 1$ convolution operations to obtain the flattened cross-domain feature representation $\mathbf{F}_{cross}\in \mathbb{R}^{d_{c}}$. Inspired by the prototype theory \cite{garcia2012prototype}, the class-wise prototype $\mathbf{P}=\left\{\mathbf{p}_{k}^{c}\right\}_{c,k=1}^{C,2}\in \mathbb{R}^{d_{c}}$ is designed to express the cluster center of feature representations belonging to the $c$-th class from the $k$-th SD. Based on minimizing the distance metric between feature embeddings and prototypes, the class of each SD sample $n\in\mathbf{F}_{cross}$ can be predicted by implementing winner-take-all classification. 
\begin{equation}
\label{eq9}
    c(n)=c^{*},\mathrm{with}(c^{*},k^{*})=\mathop{\arg\min}_{(c,k)}\left\{d_{k}^{c}\right\}_{c,k=1}^{C,2}
\end{equation}
where $d_{k}^{c}=-\cos(\mathbf{F}_{cross}, \mathbf{p}_{k}^{c})$ is defined as the negative cosine similarity metric. The probability value $p_{n}^{c}$ of SD sample $n$ belonging to $c$-th class is calculated as:
\begin{equation}
\label{eq10}
    p_{n}^{c}=\frac{{\rm exp}(-d^{c})}{\sum_{c=1}^{C}{\rm exp}(-d^{c})}
\end{equation}
where $d^{c}=\min(d_{1}^{c},d_{2}^{c})$ represents the similarity to the closest prototypes of class for the optimal SD. In this process, the distance characteristics of each class from each domain is captured and clustered without introducing extra learnable parameters. The standard CE loss is adopted to train the cross-domain feature flow as follows:
\begin{equation}
\label{eq11}
    L_{PCE}=-\frac{1}{N_{S}}\sum_{n=1}^{Ns}\sum_{c=1}^{C}\mathbf{y}_{n}\log(p_{n}^{c})
\end{equation}

\subsubsection{Intra-domain Intra-class Compactness Constraint}
Each remote sensing data source has exclusive class characteristics. To efficiently grasp different contributions of each domain, the intra-domain intra-class compactness constraint is introduced to model high-order intra-class relationships and amplify unique discrepancy characteristics to improve the generalization capability. Firstly, as illustrated in Fig. \ref{fig: CDID} (b), an intra-domain feature fusion strategy is adopted to extract feature representation based on different channel settings, which is defined as:
\begin{equation}
\label{eq12}
\begin{aligned}
    \mathbf{O}_{k} = \left\{
    \begin{array}{ll}
    \mathcal{F}({\rm Concat}(\mathbf{F}_{k}^{spa},\mathbf{F}_{k}^{cha})), & N_{k} \geq N_{k'} \\
    {\rm Concat}(\mathcal{W}(\mathbf{F}_{k}^{spa}),\mathcal{W}(\mathbf{F}_{k}^{cha})), & N_{k} \textless N_{k'} \\
    \end{array}
    \right.
\end{aligned}
\end{equation}
where $\mathcal{F}$ means multiple convolution operations to extract intra-domain feature representations. The obtained domain features $\mathbf{O}_{k}$ and $\mathbf{O}_{k'}$ are further concatenated as $\mathbf{F}_{intra}$ to send the kernel mixing module (KMM) to model high-order intra-class relationships. KMM considers high-order statistics of intra-domain features within the Gaussian Reproducing Kernel Hilbert space $\mathcal{H}$ and expresses each class-wise representation by learning three learnable kernel components: mixture coefficient $\boldsymbol{\pi}_{c}$, mean vector $\boldsymbol{\mu}_{c}$ and covariance matrix $\boldsymbol{\Sigma}_{c}$. Given an intra-domain feature sample $\mathbf{f}^{n}$ in $\mathbf{F}_{intra}$, the intra-class representation is modeled in KMM using the following expression: 
\begin{equation}
\label{eq13}
    \phi_{c}(\mathbf{f}^{n})=\pi_{c}^{n}\zeta_{c}(\mathbf{f}^{n})=\pi_{c}^{n} {\rm exp}(-\frac{\Vert\mathbf{f}^{n}-\mu_{c}^{n}\mathbf{1}\Vert^{2}}{2(\sigma_{c}^{n})^{2}})
\end{equation}
where $\zeta_{c}(\cdot)$ is an isotropic Gaussian kernel satisfying $\boldsymbol{\mu}_{c}^{n}=\mu_{c}^{n}\mathbf{1}$, $\boldsymbol{\sigma}_{c}^{n}=(\sigma_{c}^{n})^{2}\mathbf{I}$, and $\pi_{c}^{n}\in [0,1]$. $\mathbf{I}$ represents the identity matrix and $\mathbf{1}=[1,\cdots,1]^{T}$ is the one vector. $\mu_{c}^{n}$ and $\sigma_{c}^{n}$ denote the mean and covariance of the $c$-th kernel, respectively. In particular, we adopt different activation functions to satisfy constraints of these three parameters $\mathbf{\theta}_{c}^{n}=\left\{\mu_{c}^{n},\sigma_{c}^{n},\pi_{c}^{n}\right\}$. Specifically, a normal linear layer is used for $\mu_{c}^{n}$ and modified exponential linear unit (ELU) \cite{clevert2015fast} is adopted to ensure that $\sigma_{c}^{n}$ is positive. The sigmoid function is used for $\pi_{c}^{n}$ to generate the mixture coefficient in the range of $[0,1]$. The training process of KMM includes the reconstruction loss and asymmetric loss to learn high-order class-wise representation and balance the probabilities of different samples.
\begin{equation}
\label{eq14}
    L_{K}=\frac{1}{N_{S}}\sum_{n=1}^{N_{S}}\sum_{c=1}^{C}-\log(\frac{\phi_{c}(\mathbf{f}^{n})}{\pi_{c}^{n}\mathbf{y}_{n}})-\mathbf{y}_{n}L_{+}-(1-\mathbf{y}_{n})L_{-}
\end{equation}
where $L_{+}=(1-\pi_{c}^{n})^{\gamma_{+}}\log(\pi_{c}^{n})$ and $L_{-}=(\pi_{c}^{n})^{\gamma_{-}}\log(1-\pi_{c}^{n})$ represent  the positive and negative loss parts, respectively. $\gamma_{+}$ and $\gamma_{-}$ are the hyperparameters to control the contribution of positive and negative samples.

\subsection{Joint Classification of Original and Augmented MS Images}
In MS-CDG, the adversary neural network and multi-level diversification model are optimized separately. Firstly, the multi-level diversification model is pre-trained by the aggregation of cross-domain and intra-domain modeling based on original MS images, formulated as
\begin{equation}
\label{eq15}
    L_{pre}=L_{PCE}+\alpha_{1} L_{K}
\end{equation}
where $\alpha_{1}$ is a balanced hyperparameter to adjust the cross-domain and intra-domain modeling. Based on the obtained predicted probability output $\mathbf{p}_{n}^{g}$, the adversary neural network is optimized to generate the augmented MS images by $L_{ADV}$. Then, original and augmented MS images are jointly trained to increase model diversity by introducing a distribution consistency loss $L_{consist}$ as:
\begin{equation}
\label{eq16}
    L_{consist}={\rm KL}(p_{mix}|| p_{n}^{c}) + {\rm KL}(p_{mix} || p_{n}^{g})
\end{equation}
where ${\rm KL}(\cdot)$ is the Kullback-Leibler (KL) divergence \cite{kullback1951information}. $p_{mix}=(p_{n}^{g}+p_{n}^{c})/2$ denotes the mix prediction of original and augmented MS images designed to ensure class distribution alignment. Finally, the joint classification loss of MS-CDG is given by
\begin{equation}
\label{eq17}
    L_{joint}=\alpha_{2}L_{pre}+(1-\alpha_{2}) L_{consist}
\end{equation}
where $\alpha_{2}$ is the regularization parameter to balance different objective functions. The main steps of the proposed MS-CDG are summarized in \textbf{Algorithm \ref{alg:MSCDG}}. Parameters in the adversary neural network are updated at the fixed epoch intervals $T_{adv}$ to naturally calibrate the augmented MS images and enables MS-CDG to achieve higher generalization performance in the unseeen TD.  

\begin{algorithm}[!t]
\caption{Pseudo-code of the MS-CDG framework}
\label{alg:MSCDG}
{\bf Input:}
Source domain dataset $\mathcal{X}$, Label $\mathbf{y}$, Adversarial neural network $G_{adv}$, Number of maximum iteration $T$\\
{\bf Output:}
Learned task model $M$
\begin{algorithmic}[1]
    \State $\bf Initialize:$ $\mathcal{X}_{all}\gets \left\{\mathcal{X},\mathbf{y}\right\}$, weights of $G_{adv}$ randomly
    \For{$t$=1,2,\dots,$T$}
        \State Sample a mini-batch $(\mathbf{x}_{n}^{k},\mathbf{x}_{n}^{k'})$ from $\mathcal{X}$
        \If {$t<T_{pre}$}
            \State Pre-train $M$ using Eq.(\ref{eq15})
        \Else
            \For{$t$=1,2,\dots,$T_{adv}$}
            \State Train $G_{adv}$ using Eq.(\ref{eq2}) and fix $M$
            \EndFor
            \State Generate $(\mathbf{\hat{x}}_{n}^{k},\mathbf{\hat{x}}_{n}^{k'})$ from $G_{adv}$
            \State Train $M$ using Eq.(\ref{eq17}) and fix $G_{adv}$
        \EndIf
    \EndFor
    \State \Return $M$
\end{algorithmic}
\end{algorithm}

\begin{figure*}[!t]
	  \centering
		\subfigure{
			\includegraphics[width=1\textwidth]{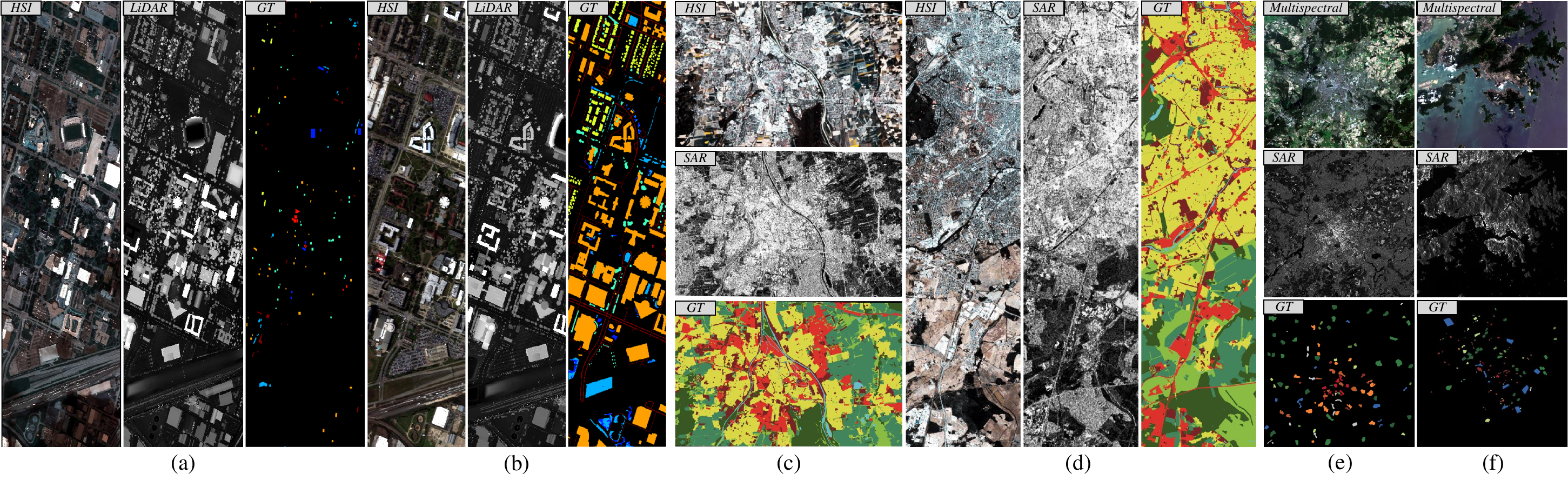}
		}
        \caption{MS remote sensing datasets, including two remote sensing data sources and the corresponding ground-truth (GT) map. (a) Houston 2013 dataset. (b) Houston 2018 dataset. (c) Augsburg dataset. (d) Berlin dataset. (e) LCZ Berlin dataset. (f) LCZ Hong Kong dataset.}
\label{fig:dataset}
\end{figure*}

\section{Experimental Results and Analysis}
In this section, eight mainstream cross-scene classification approaches based on DA and DG are adopted to conduct comparative experiments on three types of MS remote sensing datasets, including generative adversarial network (GAN) \cite{nirmal2020open}, DAAN \cite{yu2019transfer}, DSAN \cite{zhu2020deep}, TSTNet \cite{zhang2021topological}, MDA-Net \cite{zhang2023cross}, progressive domain expansion network (PDEN) \cite{li2021progressive}, SDENet \cite{zhang2023single} and LLURNet \cite{10268956}. The overall accuracy (OA), average accuracy (AA) and Kappa coefficient (Kappa) are employed to evaluate the cross-scene classification performance. Parameter tuning and ablation studies are finally conducted to provide suitable parameter settings and verify the effectiveness of individual components in MS-CDG.

\begin{table}[!t]
\centering
\caption{Number of Source and Target Samples for the Houston Datasets}
\begin{tabular}{c|c|c|c}
\hline \hline
\multicolumn{2}{c|}{Class} & \multicolumn{2}{c}{Number of Samples}\\
\hline
No. & Name & \thead{Houston 2013 \\ (Source)} & \thead{Houston 2018 \\ (Target)}\\
\hline
1 & Grass Healthy & 506 & 1574\\
2 & Grass Stressed & 492 & 5188\\
3 & Trees & 534 & 2970\\
4 & Residential Buildings & 250 & 6355\\
5 & Non-Residential Buildings & 497 & 35758\\
6 & Water & 260 & 40\\
7 & Road & 516 & 7285\\
\hline
\multicolumn{2}{c|}{Total} & 3055 & 59170\\
\hline \hline
\end{tabular}
\label{tab:houston dataset}
\end{table}

\begin{table}[!t]
\centering
\caption{Number of Source and Target Samples for the Germany Datasets}
\begin{tabular}{c|c|c|c}
\hline \hline
\multicolumn{2}{c|}{Class} & \multicolumn{2}{c}{Number of Samples}\\
\hline
No. & Name & \thead{Augsburg \\ (Source)} & \thead{Berlin \\ (Target)}\\
\hline
1 & Water & 1130 & 16651\\
2 & Residential Area & 18682 & 589409\\
3 & Industrial Area & 7546 & 171957\\
4 & Artificial Area & 6541 & 146678\\
5 & Agriculture & 14280 & 171657\\
6 & Low Plants & 5611 & 194536\\
7 & Forest & 6193 & 189305\\
\hline
\multicolumn{2}{c|}{Total} & 59983 & 1480193\\
\hline \hline
\end{tabular}
\label{tab:germany dataset}
\end{table}

\begin{table}[!t]
\centering
\caption{Number of Source and Target Samples for the LCZ Datasets}
\begin{tabular}{c|c|c|c}
\hline \hline
\multicolumn{2}{c|}{Class} & \multicolumn{2}{c}{Number of Samples}\\
\hline
No. & Name & \thead{Berlin \\ (Source)} & \thead{Hong Kong \\ (Target)}\\
\hline
1 & Compact Mid-rise & 153 & 179\\
2 & Open High-rise & 58 & 673\\
3 & Open Mid-rise & 245 & 126\\
4 & Open Low-rise & 401 & 120\\
5 & Large Low-rise & 165 & 137\\
6 & Dense Trees & 496 & 1616\\
7 & Scattered Trees & 103 & 540\\
8 & Bush and Scrub & 105 & 691\\
9 & Low Plants & 442 & 985\\
10 & Water & 173 & 2603\\
\hline
\multicolumn{2}{c|}{Total} & 2341 & 7670\\
\hline \hline
\end{tabular}
\label{tab:lcz dataset}
\end{table}

\subsection{Data Description and Experimental Setup}
\subsubsection{Houston Datasets}
The Houston datasets include two sets of MS remote sensing data across time scales, namely Houston 2013 \cite{debes2014hyperspectral} and Houston 2018 \cite{le20182018} acquired by different sensors over the University of Houston and its neighboring area. Each data consists of HSI and LiDAR-derived digital surface model (DSM). These datasets were released in 2013 and 2018 by IEEE GRSS DFC\footnote{https://hyperspectral.ee.uh.edu/}. As illustrated in Fig. \ref{fig:dataset} (a) and (b), the overlapping and non-overlapping areas of Houston 2013 and Houston 2018 are cropped and selected as the study area. The Houston 2013 dataset is adopted as SD and has $349\times 954$ pixels with 48 spectral bands corresponding to the Houston 2018 dataset. The Houston 2018 dataset is regarded as TD and has $240\times 954$ pixels. Seven consistent classes are covered and the corresponding number of source and target samples is listed in Table \ref{tab:houston dataset}. 

\begin{table*}[!t]
\centering
\caption{Class-Specific and Overall Classification Accuracy (\%) of Different Methods on the Houston Datasets}
\begin{tabular}{c||c|c|c|c|c|c|c|c|c}
\hline \hline
\multirow{2}{*}{Class} & \multicolumn{9}{c}{Classification Algorithm}\\
\cline{2-10}
~ & GAN \cite{nirmal2020open} & DAAN \cite{yu2019transfer} & DSAN \cite{zhu2020deep} & TSTNet \cite{zhang2021topological} & MDA-Net \cite{zhang2023cross} & PDEN \cite{li2021progressive} & SDENet \cite{zhang2023single} & LLURNet \cite{10268956} & MS-CDG\\
\hline
1 & \bf 70.71 & 33.48 & 23.06 & 54.57 & 66.84 & 19.63 & 46.63 & 28.65 & 44.85\\
2 & 63.80 & 83.53 & 84.46 & 85.75 & 80.03 & 40.18 & 85.65 & \bf 86.83 & 77.35\\
3 & 58.65 & \bf 78.41 & 71.94 & 70.79 & 64.72 & 59.80 & 68.53 & 51.47 & 64.92\\
4 & 56.34 & 65.24 & 65.80 & 78.75 & 56.06 & 31.14 & 47.15 & \bf 82.75 & 66.82\\
5 & 40.11 & 72.97 & 77.15 & 78.29 & 81.07 & 82.16 & 88.31 & 81.18 & \bf 90.08\\
6 & 62.50 & 95.00 & \bf 100.00 & 90.00 & 87.50 & 40.00 & 90.00 & 97.50 & \bf 100.00\\
7 & 64.49 & 48.37 & 64.77 & 44.08 & 76.80 & 23.09 & 55.59 & 34.94 & \bf 76.84\\
\hline
OA (\%) & 48.70$\pm$1.78 & 69.29$\pm$3.05 & 73.36$\pm$2.04 & 76.53$\pm$2.92 & 76.57$\pm$2.02 & 63.32$\pm$2.16 & 77.53$\pm$1.54 & 73.28$\pm$1.59 & \bf 81.87$\pm$0.98\\
\hline
AA (\%) & 59.51$\pm$4.14 & 68.14$\pm$1.41 & 69.60$\pm$1.31 & 71.75$\pm$1.75 & 73.29$\pm$1.44 & 42.29$\pm$2.83 & 68.84$\pm$3.84 & 66.19$\pm$3.56 & \bf 74.41$\pm$1.26\\
\hline
Kappa (\%) & 33.20$\pm$2.32 & 54.18$\pm$3.51 & 59.56$\pm$1.92 & 61.88$\pm$3.35 & 64.03$\pm$2.56 & 39.28$\pm$2.57 & 62.49$\pm$2.83 & 56.89$\pm$2.86 & \bf 70.95$\pm$1.53\\
\hline \hline
\end{tabular}
\label{tab:houston result}
\end{table*}

\begin{table*}[!t]
\centering
\caption{Class-Specific and Overall Classification Accuracy (\%) of Different Methods on the Germany Datasets}
\begin{tabular}{c||c|c|c|c|c|c|c|c|c}
\hline \hline
\multirow{2}{*}{Class} & \multicolumn{9}{c}{Classification Algorithm}\\
\cline{2-10}
~ & GAN \cite{nirmal2020open} & DAAN \cite{yu2019transfer} & DSAN \cite{zhu2020deep} & TSTNet \cite{zhang2021topological} & MDA-Net \cite{zhang2023cross} & PDEN \cite{li2021progressive} & SDENet \cite{zhang2023single} & LLURNet \cite{10268956} & MS-CDG\\
\hline
1 & 22.48 & 18.11 & \bf 30.67 & 15.34 & 16.50 & 5.58 & 13.93 & 10.59 & 15.67\\
2 & 61.11 & 57.65 & 39.81 & 70.59 & 66.68 & \bf 96.06 & 44.25 & 72.34 & 67.10\\
3 & 36.21 & 25.22 & 40.08 & 33.41 & 27.91 & 40.25 & 36.01 & 26.54 & \bf 52.30\\
4 & 24.42 & 53.41 & 53.24 & 33.63 & 49.41 & 11.29 & \bf 60.10 & 23.66 & 42.58\\
5 & 15.02 & 43.84 & \bf 64.05 & 39.11 & 36.43 & 8.03 & 32.08 & 59.15 & 58.49\\
6 & 6.84 & 8.02 & 22.95 & 13.14 & \bf 46.21 & 10.71 & 15.97 & 44.92 & 26.47\\
7 & 66.06 & 70.83 & 70.25 & \bf 80.38 & 65.43 & 41.24 & 77.39 & 50.11 & 60.29\\
\hline
OA (\%) & 42.30$\pm$1.42 & 46.57$\pm$2.46 & 45.67$\pm$2.49 & 52.04$\pm$2.08 & 53.55$\pm$1.45 & 51.74$\pm$3.05 & 43.68$\pm$2.36 & 53.53$\pm$1.99 & \bf 56.56$\pm$1.18\\
\hline
AA (\%) & 33.16$\pm$1.14 & 39.58$\pm$1.35 & 45.86$\pm$1.58 & 40.80$\pm$0.87 & 44.08$\pm$0.68 & 30.45$\pm$1.98 & 40.05$\pm$1.32 & 41.04$\pm$1.97 & \bf 46.13$\pm$0.95\\
\hline
Kappa (\%) & 23.48$\pm$1.41 & 33.31$\pm$1.69 & 34.33$\pm$2.36 & 36.67$\pm$2.52 & 40.36$\pm$1.23 & 28.61$\pm$3.10 & 31.02$\pm$2.84 & 38.74$\pm$1.29 & \bf 42.67$\pm$1.37\\
\hline \hline
\end{tabular}
\label{tab:germany result}
\end{table*}

\begin{table*}[!t]
\centering
\caption{Class-Specific and Overall Classification Accuracy (\%) of Different Methods on the LCZ Datasets}
\begin{tabular}{c||c|c|c|c|c|c|c|c|c}
\hline \hline
\multirow{2}{*}{Class} & \multicolumn{9}{c}{Classification Algorithm}\\
\cline{2-10}
~ & GAN \cite{nirmal2020open} & DAAN \cite{yu2019transfer} & DSAN \cite{zhu2020deep} & TSTNet \cite{zhang2021topological} & MDA-Net \cite{zhang2023cross} & PDEN \cite{li2021progressive} & SDENet \cite{zhang2023single} & LLURNet \cite{10268956} & MS-CDG\\
\hline
1 & 44.69 & 26.26 & 11.17 & 7.82 & 29.83 & 6.15 & 37.43 & 41.90 & \bf 69.48\\
2 & 0 & 0 & 0 & 0 & 0 & 0 & 2.38 & 0.44 & \bf 8.31\\
3 & 19.05 & 34.92 & 30.95 & 21.43 & 23.20 & \bf 50.79 & 27.78 & 30.16 & 11.11\\
4 & 45.00 & 14.17 & 29.17 & 16.67 & \bf 60.83 & 20.83 & 20.00 & 10.83 & 17.50\\
5 & 24.82 & 57.66 & 34.31 & 27.74 & 5.84 & 3.65 & 60.58 & \bf 67.88 & 51.09\\
6 & 81.00 & 87.13 & 82.30 & 93.56 & 75.80 & 85.46 & 93.63 & 90.90 & \bf 97.41\\
7 & 9.44 & 26.67 & \bf 49.63 & 38.70 & 2.22 & 47.59 & 22.22 & 20.93 & 13.70\\
8 & 12.88 & 2.32 & \bf 28.65 & 0.14 & 0.58 & 0.14 & 4.63 & 14.62 & 6.95\\
9 & 12.28 & 2.74 & 16.35 & 30.86 & \bf 44.57 & 0.91 & 0.30 & 6.29 & 19.49\\
10 & 99.27 & 76.95 & 27.47 & 33.46 & 86.94 & 98.39 & 99.31 & 98.46 & \bf 100.00\\
\hline
OA (\%) & 56.66$\pm$1.87 & 49.35$\pm$2.65 & 36.68$\pm$2.28 & 39.06$\pm$4.77 & 52.46$\pm$4.57 & 56.25$\pm$3.03 & 58.38$\pm$1.35 & 59.06$\pm$1.17 & \bf 61.77$\pm$0.81\\
\hline
AA (\%) & 34.84$\pm$0.99 & 32.88$\pm$3.26 & 31.00$\pm$1.04 & 27.04$\pm$2.88 & 32.98$\pm$3.91 & 31.39$\pm$3.59 & 36.83$\pm$2.91 & 38.24$\pm$1.56 & \bf 39.50$\pm$1.50\\
\hline
Kappa (\%) & 45.58$\pm$0.86 & 39.34$\pm$2.06 & 29.07$\pm$2.18 & 30.55$\pm$4.75 & 42.13$\pm$4.62 & 46.48$\pm$3.26 & 48.01$\pm$1.80 & 49.13$\pm$0.78 & \bf 50.88$\pm$0.76\\
\hline \hline
\end{tabular}
\label{tab:lcz result}
\end{table*}

\subsubsection{Germany Datasets}
The second datasets contain different study areas in the Germany cities of Augsburg and Berlin. The HSI and dual-polarized SAR data is acquired by EnMAP and Sentinel-1 satellite, and georeferenced in the same area \cite{rs70708830}. The resulting images have $886\times 1360$ and $2380\times 622$ pixels in the Augsburg and Berlin area, respectively. The HSI data has the same 242 spectral band in the wavelength range of 400 nm to 2500 nm, and SAR data has two channels with VV and VH. Fig. \ref{fig:dataset} (c) and (d) visualize the false-color map of the HSI and SAR data. There are seven classes involved in this dataset, and Table \ref{tab:germany dataset} lists the number of source and target samples.

\subsubsection{LCZ Datasets}
The LCZ datasets are collected from Sentinel-1 and Sentinel-2 satellites, where the former is the dual-polarization SAR data organized as a commonly used PolSAR covariance matrix (four components) and the latter is the multispectral data containing ten spectral bands. The labeled ground truth (GT) for the Berlin and Hong Kong region can be downloaded from the IEEE GRSS DFC2017\footnote{http://www.grss-ieee.org/2017-ieee-grss-data-fusion-contest/}. As displayed in Fig. \ref{fig:dataset} (e) and (f), the Berlin region with the size of $626\times 643$ is adopted as SD to train the model, while the Hong Kong region with the size of $529\times 528$ is treated as TD for testing. The detailed surface object classes and the adopted sample numbers are listed in Table \ref{tab:lcz dataset}.

For a fair comparison, all experiments in this paper are carried out with the PyTorch framework on NVIDIA GTX 1080Ti 11-GB GPU. The Adam optimizer with a learning rate of $1e-3$ is used for training the backbone. The learning rate of the adversarial neural network is set to $5e-6$. The batch size is set to 64 for the Houston and LCZ datasets and 256 for the Germany datasets. The input is set as patch size of $13\times 13$, and the default value for $\mathit{l}_{2}$-norm regularization of all modules is set to $1e-4$ for weight decay to avoid overfitting problems. The output dimension $d_{spa}$ of encoded spatial features $\mathbf{F}_{k}^{spa}$ are empirically set to 32 for the Houston and LCZ datasets, and 64 for the Germany datasets. The output dimension $d_{cha}$ of encoded channel features $\mathbf{F}_{k}^{cha}$ is fixed at 3. The hyperparameters of $\gamma_{+}$ and $\gamma_{-}$ are set to 2 and 4 on all public datasets. The learning process will be stopped after 500 epochs. The number of pre-trained epochs $T_{pre}$ and the fixed epoch intervals $T_{adv}$ are set to 2 and 10, respectively.

\begin{figure*}[!t]
	  \centering
		\subfigure{
			\includegraphics[width=1\textwidth]{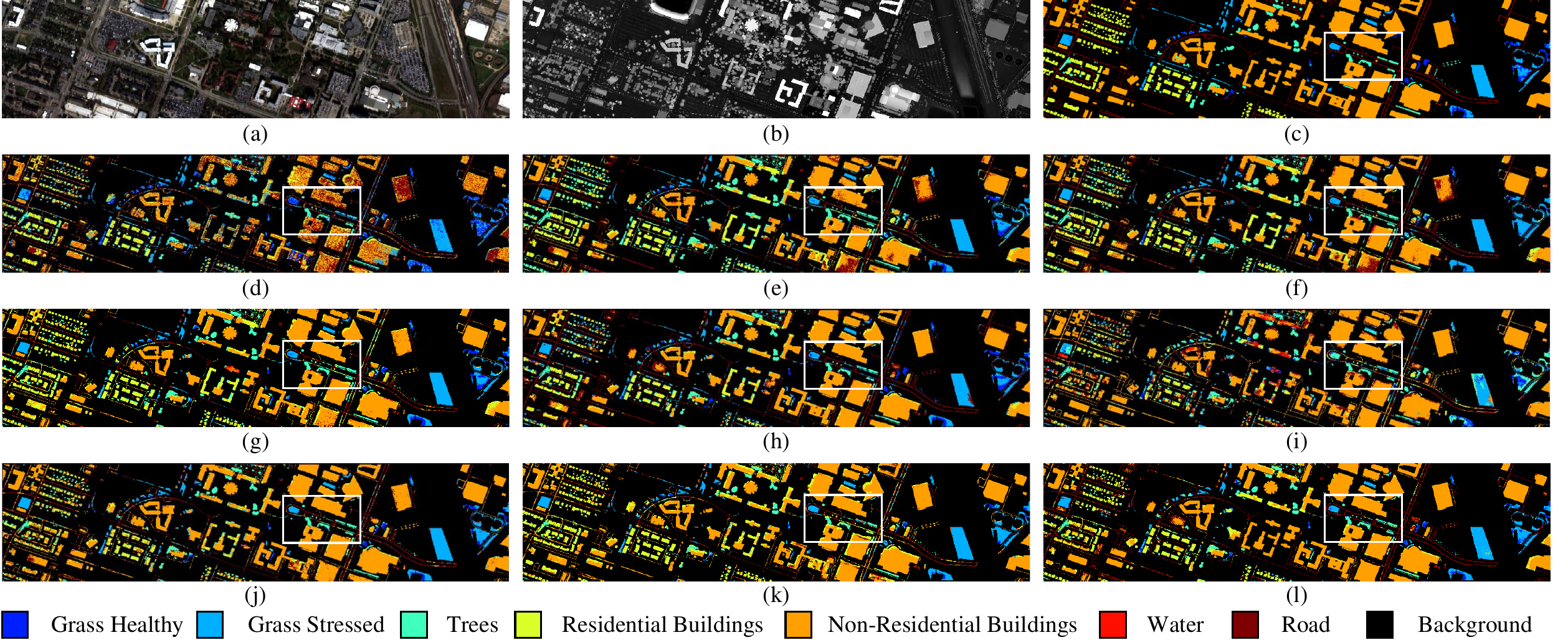}
		}
        \caption{Visualization results of different compared methods on the Houston datasets. (a) pseudocolor HSI. (b) LiDAR image. (c) Ground-truth map. (d) GAN (48.70\%). (e) DAAN (69.29\%). (f) DSAN (73.36\%). (g) TSTNet (76.53\%). (h) MDA-Net (76.57\%). (i) PDEN (63.32\%). (j) SDENet (77.53\%). (k) LLURNet (73.28\%). (l) MS-CDG (81.87\%).}
\label{fig:houston fig}
\end{figure*}

\begin{figure*}[!t]
	  \centering
		\subfigure{
			\includegraphics[width=1\textwidth]{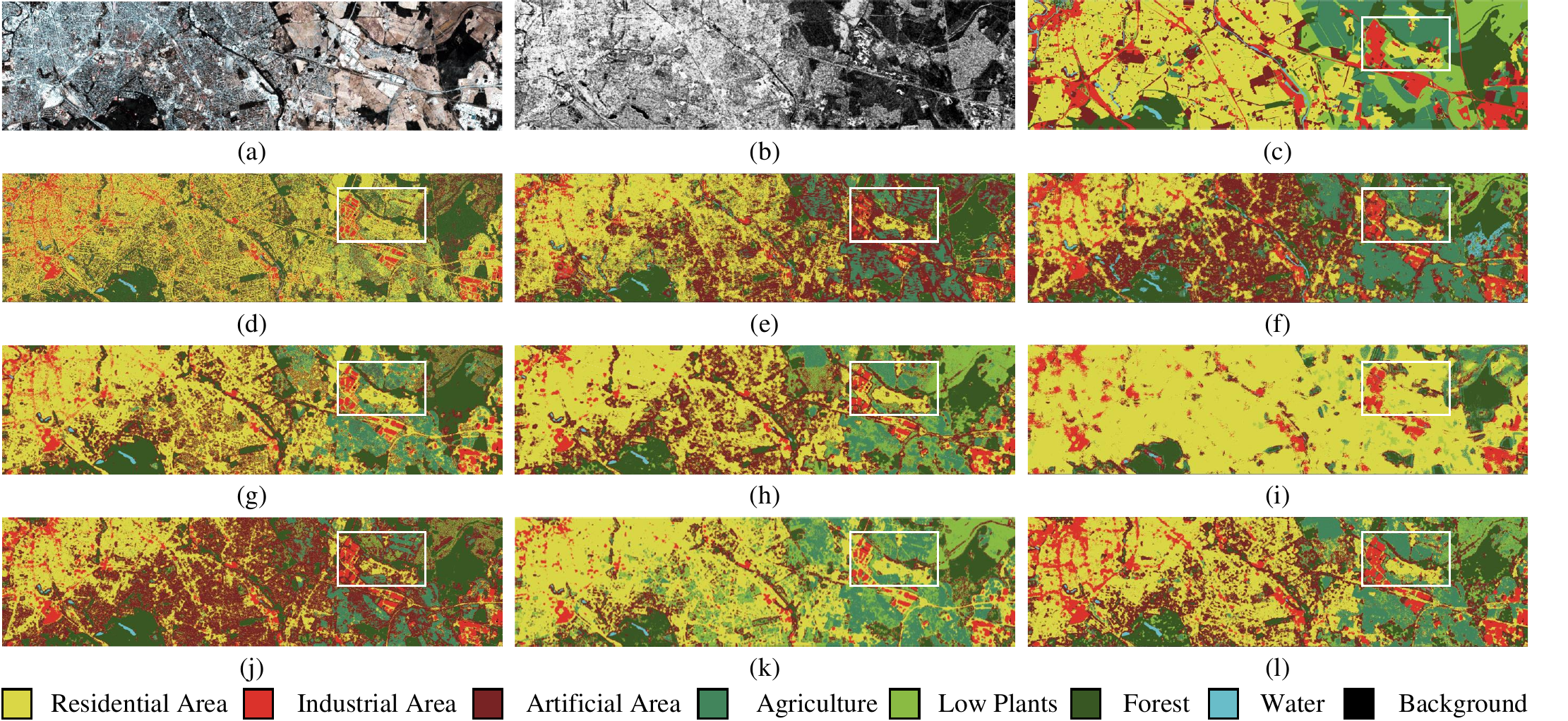}
		}
        \caption{Visualization results of different compared methods on the Germany datasets. (a) pseudocolor HSI. (b) SAR image. (c) Ground-truth map. (d) GAN (42.30\%). (e) DAAN (46.57\%). (f) DSAN (45.67\%). (g) TSTNet (52.04\%). (h) MDA-Net (53.55\%). (i) PDEN (51.74\%). (j) SDENet (43.68\%). (k) LLURNet (53.53\%). (l) MS-CDG (56.56\%).}
\label{fig:germany fig}
\end{figure*}

\begin{figure}[!t]
	  \centering
		\subfigure{
			\includegraphics[width=0.5\textwidth]{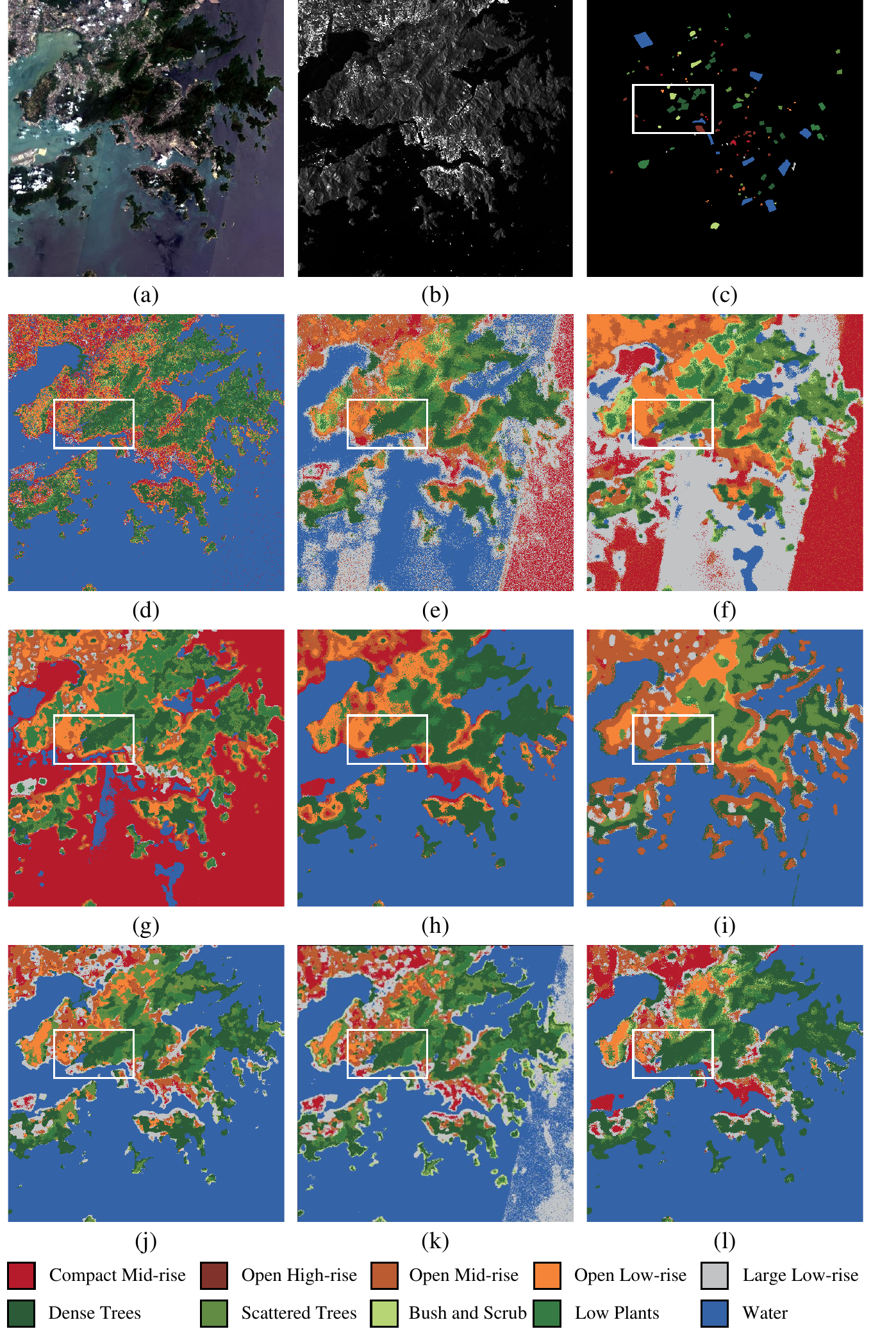}
		}
        \caption{Visualization results of different compared methods on the LCZ datasets. (a) pseudocolor multispectral image. (b) SAR image. (c) Ground-truth map. (d) GAN (56.66\%). (e) DAAN (49.35\%). (f) DSAN (36.68\%). (g) TSTNet (39.06\%). (h) MDA-Net (52.46\%). (i) PDEN (56.25\%). (j) SDENet (58.38\%). (k) LLURNet (59.06\%). (l) MS-CDG (61.77\%).}
\label{fig:lcz fig}
\end{figure}

\begin{figure}[!t]
	  \centering
		\subfigure{
			\includegraphics[width=0.5\textwidth]{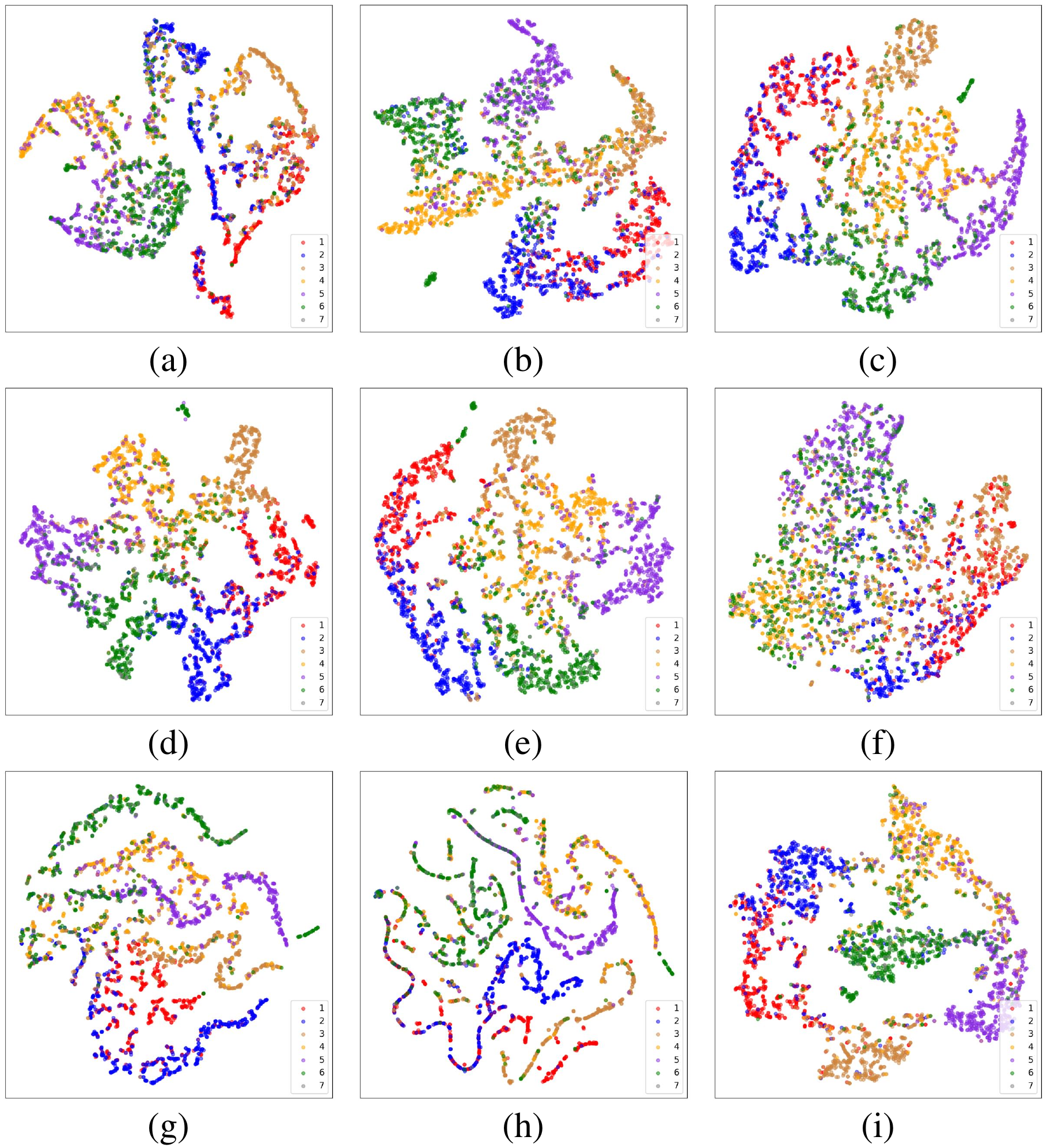}
		}
        \caption{t-SNE embedding visualization of different cross-scene classification methods on the Houston datasets. (a) GAN. (b) DAAN. (c) DSAN. (d) TSTNet. (e) MDA-Net. (f) PDEN. (g) SDENet. (h) LLURNet. (i) MS-CDG.}
\label{fig:tsne}
\end{figure}

\subsection{Comparison with Current Methods}
We conduct comparative experiments against eight state-of-the-art cross-scene classification models on three types of MS datasets. All the competing approaches and the proposed method are trained without the utilization of real labels in the TD, which ensures the fairness of the experiment. More specifically, GAN, DAAN, DSAN, TSTNet, and MDA-Net are regarded as the DA-based methods, and all TD samples without labels are used for training. For the DG-based comparison methods, PDEN, SDENet and LLURNet only adopt SD samples with labels as the training data. The mean values and standard deviations of ten independent repeated experiments are calculated for different cross-scene classification methods.

Tables \ref{tab:houston result}-\ref{tab:lcz result} report the quantitative analysis results of different methods on Houston, Germany and LCZ datasets. The corresponding visualization results of classification maps are shown in Figs. \ref{fig:houston fig}-\ref{fig:lcz fig}. The white rectangle circles indicate the areas that MS-CDG classifies more accurately compared with the comparison algorithm. In the comparison of DA-based methods, MDA-Net performs well on Houston and Germany datasets, but produces a large amount of misclassification on LCZ datasets due to the biased guidance from the high proportion of classes in the MS data. Despite the introduction of MS data, MDA-Net lacks effective class modeling and cannot achieve better classification performance in the scenario with uneven distribution of SD and TD classes. GAN can obtain better classification results than MDA-Net by an approximately 4\% increment on the LCZ datasets, which reflects that the adversarial loss can effectively deal with the impact of classifier bias in the case of imbalanced sample proportions. Among the DG-based comparison approaches, LLURNet exhibits better generalization performance by introducing locally information modeling on Germany and LCZ datasets, while SDENet performs well on Houston datasets owing to its more suitable generation strategy for scenes with small spatial layout changes. With the help of data-aware adversarial augmentation and model-aware multi-level diversification, it can be observed that MS-CDG is improved by 2\% to 5\% over MDA-Net and LLURNet on all TDs. Due to large differences of label distributions on LCZ datasets located in complex urban areas, the 2-nd class (Open High-rise) is easy to be wrongly classified as the 4-th class (Open Low-rise) in all methods, but MS-CDG can still identify a few class features. The classification maps are closer to GT and the accuracy of specific classes is greatly improved, such as the  5-th (Non-Residential Buildings) and 7-th (Road) classes in Fig. \ref{fig:houston fig}, 3-th (Industrial Area) class in Fig. \ref{fig:germany fig}, and 1-st (Compact Mid-rise) and 6-th (Dense Trees) classes in Fig. \ref{fig:lcz fig}. Overall, the proposed MS-CDG can surpass all comparison methods and obtain the best classification performance on three types of MS datasets. The extracted domain-invariant features enable the proposed MS-CDG to develop a more profound comprehension of the input MS data and enhance cross-scene classification performance.

\begin{figure}[!t]
	  \centering
		\subfigure{
			\includegraphics[width=0.5\textwidth]{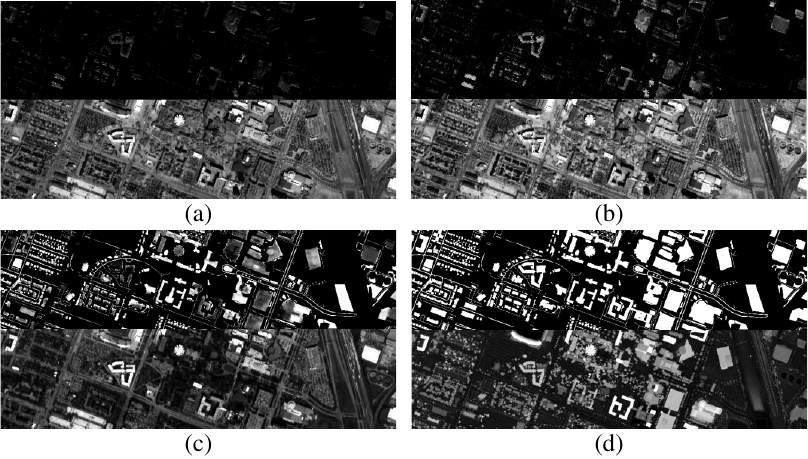}
		}
        \caption{Visualization results of the augmented and original MS image on the Houston datasets. (a) HSI in the 10-th band. (b) HSI in the 20-th band. (c) HSI in the 30-th band. (d) LiDAR image.}
\label{fig:generate}
\end{figure}

The t-distributed stochastic neighbor embedding (t-SNE) \cite{van2008visualizing} is adopted to produce better visualizations for investigating the class separability performance of different cross-scene classification methods on the Houston datasets, as shown in Fig. \ref{fig:tsne}. Note that, the adopted embedding features are output by the last network layer, which represents the learned class relationship in different approaches. It is apparent that the inter-class distributions of some DA approaches are still mixed up, e.g., DSAN and TSTNet. The main reason is that the DA-based methods focus on the alignment of SD and TD features by reducing domain shifts, and ignore the specific class modeling within each domain. In addition, existing modeling strategies, such as subdomain adaptation or topological techniques, cannot satisfy the domain differences of multiple classes, and different classes are mapped into similar feature spaces, further exacerbating the difficulty of class separation. The introduction of MS data helps MDA-Net improve a certain intra-class aggregation, such as the 1-st and 2-nd class, but the confounding phenomenon of different classes still exists. Since the prior knowledge of TD is unknown, the DG-based methods rely on the extracted domain-invariant representation to generalize TD. In this process, the latent association of classes are effectively learned, so that the specific class with lower prior proportions can be better distinguished compared with the DA-based methods, e.g., the 5-th class. Overall, the proposed MS-CDG achieves higher separability for different classes than other baselines, demonstrating the effectiveness of class-specific modeling that considers both cross-domain and intra-domain feature spaces. Fig. \ref{fig:generate} illustrates the visualization results of the augmented and original MS image on the Houston datasets. It can be seen that MS-CDG can generate MS images that reflects various ground object characteristics in different bands. Through the design of the adversary neural network, MS-CDG can maintain a more significant semantic structure in the GT distribution, further helping the model achieve effective joint classification.

\begin{table*}[!t]
    \centering
    \caption{The execution time (in seconds) of one epoch training and inference in different cross-scene classification methods.}
    \resizebox{1\textwidth}{!}{
    \begin{tabular}{ccccccccccc}
        \hline \hline
       ~ & Methods & GAN \cite{nirmal2020open} & DAAN \cite{yu2019transfer} & DSAN \cite{zhu2020deep} & TSTNet \cite{zhang2021topological} & MDA-Net \cite{zhang2023cross} & PDEN \cite{li2021progressive} & SDENet \cite{zhang2023single} & LLURNet \cite{10268956} & MS-CDG\\
         \hline
         \multirow{2}{*}{Houston} & Training & 22.97 & 166.81 & 172.90 & 271.06 & 391.52 & 76.24 & 71.58 & 56.25 & \bf 4.51\\
         ~ & Inference & 7.15 & 28.13 & 25.72 & 13.08 & 13.78 & 61.47 & 12.68 & 12.67 & \bf 3.83\\
         \hline
         \multirow{2}{*}{Germany} & Training & 82.55 & 247.93 & 292.31 & 279.98 & 372.94 & 153.24 & 125.77 & 96.70 & \bf 75.24\\
         ~ & Inference & 93.98 & 211.67 & 266.70 & 308.62 & 342.27 & 342.24 & 125.29 & 96.27 & \bf 87.86\\
         \hline
         \multirow{2}{*}{LCZ} & Training & 3.87 & 9.25 & 11.05 & 17.63 & 28.36 & 5.81 & 5.27 & 6.13 & \bf 3.28\\
         ~ & Inference & 0.79 & 1.83 & 1.05 & 1.45 & 2.16 & 3.19 & 0.88 & 0.87 & \bf 0.67\\
        \hline \hline
    \end{tabular}
    }
    \label{tab:computation}
\end{table*}

\begin{figure*}[!t]
	  \centering
		\subfigure{
			\includegraphics[width=0.95\textwidth]{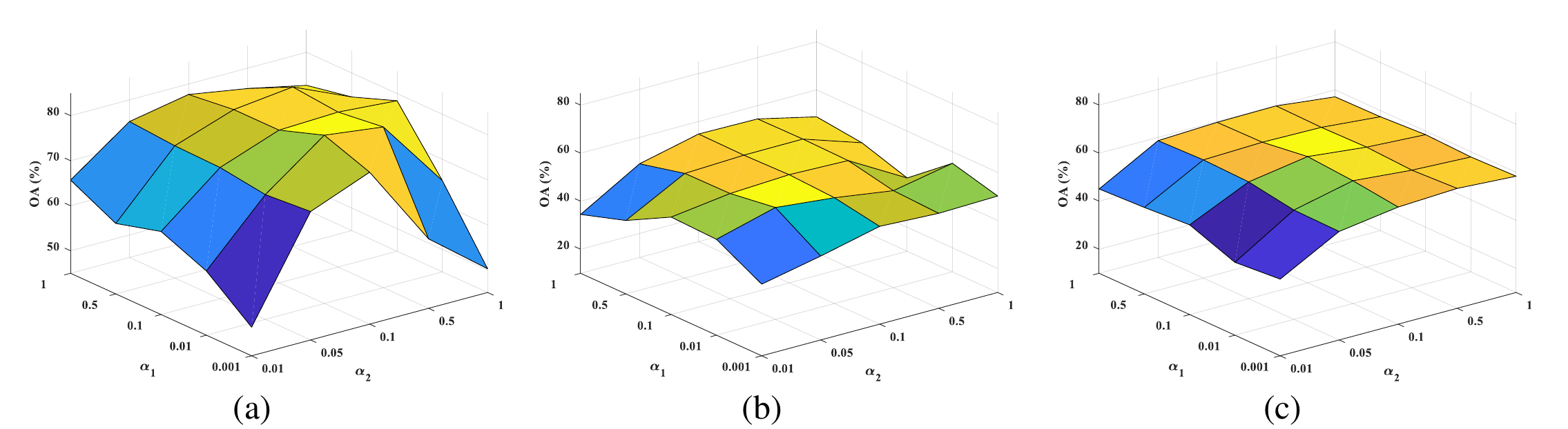}
		}
        \caption{Parameter tuning of $\alpha_{1}$ and $\alpha_{2}$ for the proposed MS-CDG using all the three experimental datasets. (a) Houston. (b) Germany. (c) LCZ.}
\label{fig:lambda}
\end{figure*}

To display computational complexity of different cross-scene classification methods, the one epoch training time and inference time on three datasets are listed in Table \ref{tab:computation}. For the DA-based methods, the joint learning of SD and TD usually consumes more time to align different domain features than the DG-based methods except GAN. The continuous training between the generator and classifier helps GAN obtain high accuracy precision with less data samples. In addition, the introduction of topological architecture and MS data increase the calculation time due to convergence difficulty during the training process, such as TSTNet and MDA-Net. PDEN has the worst inference time because it takes more time to run multiple domain generators and combine their respective results during the inference phase. Taken together, it is apparent that MS-CDG can obtain lower computational cost than other comparison methods on all datasets. The double-branch network design for cross-domain and intra-domain class modeling ensures better domain-invariant representation learning without bringing additional computational burden to the entire model.

\subsection{Parameter Tuning}
In this section, the parameter sensitivity analysis is conducted for the regularization parameters $(\alpha_{1}, \alpha_{2})$ and the number of layers in the adversarial neural network. The adjustable ranges of $\alpha_{1}$ and $\alpha_{2}$ are set as $\left\{0.001, 0.01, 0.1, 0.5, 1\right\}$ and $\left\{0.01, 0.05, 0.1, 0.5, 1\right\}$, respectively. Besides, the values of the number of layers in adversarial neural network are selected from $\left\{2, 3, 4, 5, 6\right\}$, which reflects the degree of parameter sharing under different layer settings and also affect the quality of the augmented MS samples. For instance, there is no shared convolution layer $T_{shared}$ when the number of layers is 2, and only independent convolution and residual structures are included. The number of layers is 3, meaning that only one layer is shared for different SDs and the other two layers are adopted to transform the input MS data.

Fig. \ref{fig:lambda} provides OA results of MS-CDG corresponding to various combinations of $\alpha_{1}$ and $\alpha_{2}$ on three types of MS remote sensing datasets. It can be seen from Fig. \ref{fig:lambda} that the proposed method with parameter $\alpha_{2}$ in the range of $0.05$ to $0.5$ can achieve sub-optimal classification performance, and $\alpha_{1}$ has its own sub-optimal range in different datasets. Consequently, the optimal parameters of $\alpha_{1}$ and $\alpha_{2}$ are listed as follows: $\alpha_{1}=0.01$ and $\alpha_{2}=0.1$ for Houston datasets, $\alpha_{1}=0.01$ and $\alpha_{2}=0.05$ for Germany datasets, $\alpha_{1}=0.1$ and $\alpha_{2}=0.1$ for LCZ datasets. Furthermore, the sensitivity analysis of the number of layers is performed and Table \ref{tab:adv layer} presents the cross-scene classification performance comparison on all experimental datasets. It can be observed that the classification performance is unsatisfactory when there is no shared layer, because the augmented MS samples lack the guidance of feature attribute interaction in different SDs and further affect the model training. The introduction of shared layers in the adversarial neural network can promote the model's comprehensive integration of domain style and semantic information, so that the augmented MS samples are more reliable. Due to band differences of different SDs, the optimal number of layers in this paper is set to 5 for Houston and Germany datasets, and 4 for LCZ datasets. 

\begin{table}[!t]
    \centering
    \caption{Comparison of Cross-scene Classification for Different Layers in the Adversarial Neural Network.}
    \resizebox{0.45\textwidth}{!}{
    \begin{tabular}{c|c|c|c|c|c|c}
        \hline \hline
        \multirow{2}{*}{Datasets} & \multirow{2}{*}{Metrics} & \multicolumn{4}{c}{Number of Layers}\\
        \cline{3-7}
         & & 2 & 3 & 4 & 5 & 6 \\
         \hline
         \multirow{3}{*}{Houston} & OA (\%) & 77.67 & 77.82 & 80.3 & \bf 81.87 & 80.69\\
         & AA (\%) & 67.25 & 69.95 & 70.68 & \bf 74.41 & 72.38\\
         & Kappa (\%) & 63.87 & 64.49 & 69.22 & \bf 70.95 & 69.57\\
         \hline
         \multirow{3}{*}{Germany} & OA (\%) & 47.82 & 50.39 & 55.18 & \bf 56.56 & 54.64\\
         & AA (\%) & 40.42 & 37.52 & 43.99 & \bf 46.13 & 44.79\\
         & Kappa (\%) & 33.69 & 37.94 & 41.94 & \bf 42.67 & 41.01\\
         \hline
         \multirow{3}{*}{LCZ} & OA (\%) & 56.57 & 59.39 & \bf 61.77 & 61.11 & 58.59\\
         & AA (\%) & 36.60 & 37.46 & \bf 39.50 & 38.85 & 36.87\\
         & Kappa (\%) & 45.81 & 49.35 & \bf 50.88 & 50.29 & 48.62\\
         \hline
        \hline \hline
    \end{tabular}
    }
    \label{tab:adv layer}
\end{table}

\subsection{Ablation Study}
To assess the contribution of individual elements in the proposed MS-CDG, the ablation experiment is conducted in this section by incorporating additional components to the baseline. There are four variants in the ablation analysis, (1) $L_{PCE}$: the class-wise prototype clustering is applied in the cross-domain feature flow to reduce domain discrepancies, (2) $L_{K}$: the intra-class compactness constraint is adopted for the extracted intra-domain features to aggregate high-dimensional class correlations, (3) $L_{ADV}$: the adversarial neural network is trained by introducing the learned semantic guide to enhance the quality of MS data augmentation, (4) $L_{consist}$: the consistency constraint is achieved to ensure class distribution alignment between original and augmented MS data. As shown in Table \ref{tab:ablation}, it can be seen that the proposed MS-CDG outperforms other variants and yields the highest precision on three public MS remote sensing datasets. More specifically, OA, AA, and Kappa exceed the baseline (only $L_{PCE}$) by 5.42\%, 7.88\%, 7.36\% on Houston datasets, respectively. On Germany and LCZ datasets, the increases are 7.55\%, 4.05\%, 6.65\%, and 3.91\%, 3.74\% and 3.29\%, respectively. The performance of introducing intra-class compactness constraint based on KMM can effectively improve the cross-scene classification performance, indicating that the combination of cross-domain and intra-domain class modeling plays an important role in extracting domain-invariant representations. Feeding the augmented samples with semantic information into model training can significantly improve classification performance from 1\% to 5\%. Furthermore, the class distribution alignment can enhance the overall performance by approximately 1\% in OA on all datasets. Compared with OA and Kappa, the proposed method has a more significant improvement on AA, because the joint classification of original images and augmented images is conducive to learning class modeling and enhancing class separability. Although the fluctuation in the accuracy of Germany datasets is smaller than other two datasets owing to the excessive number of labeled samples, it is evident that the consistency distribution of the augmented and original samples in MS-CDG can increase the diversity of training samples and achieve an overall enhancement in cross-scene classification performance. Overall, no matter from the data level or the model level, each component of MS-CDG can boost the cross-scene classification performance in three evaluation indicators, which fully verifies the effectiveness of the data-aware adversarial augmentation module and model-aware multi-level diversification module. 

\begin{table}[!t]
    \centering
    \caption{Ablation comparison of each variant in MS-CDG.}
    \resizebox{0.5\textwidth}{!}{
    \begin{tabular}{cccc|c|c|c|c}
        \hline \hline
        \multicolumn{4}{c|}{Module} & \multirow{2}{*}{Metrics} & \multicolumn{3}{c}{Datasets}\\
        \cline{1-4} \cline{6-8}
         $L_{PCE}$ & $L_{K}$ & $L_{ADV}$ & $L_{consist}$ & & Houston & Germany & LCZ \\
         \hline
        \multirow{3}{*}{\cmark} & \multirow{3}{*}{\xmark} & \multirow{3}{*}{\xmark} & \multirow{3}{*}{\xmark} & OA (\%) & 76.45 & 49.01 & 57.86\\
        & & & & AA (\%) & 66.53 & 42.08 & 35.76\\
        & & & & Kappa (\%) & 63.59 & 36.02 & 47.59\\
        \hline
        \multirow{3}{*}{\cmark} & \multirow{3}{*}{\cmark} & \multirow{3}{*}{\xmark} & \multirow{3}{*}{\xmark} & OA (\%) & 78.77 & 54.93 & 59.29\\
        & & & & AA (\%) & 68.90 & 44.75 & 37.61\\
        & & & & Kappa (\%) & 66.58 & 41.79 & 49.12\\
        \hline
        \multirow{3}{*}{\cmark} & \multirow{3}{*}{\xmark} & \multirow{3}{*}{\cmark} & \multirow{3}{*}{\xmark} & OA (\%) & 79.01 & 54.72 & 58.70\\
        & & & & AA (\%) & 69.51 & 43.96 & 36.63\\
        & & & & Kappa (\%) & 68.50 & 41.22 & 48.38\\
        \hline
        \multirow{3}{*}{\cmark} & \multirow{3}{*}{\xmark} & \multirow{3}{*}{\cmark} & \multirow{3}{*}{\cmark} & OA (\%) & 79.65 & 55.84 & 59.43\\
        & & & & AA (\%) & 70.75 & 45.02 & 38.91\\
        & & & & Kappa (\%) & 69.16 & 42.15 & 49.56\\
        \hline
        \multirow{3}{*}{\cmark} & \multirow{3}{*}{\cmark} & \multirow{3}{*}{\cmark} & \multirow{3}{*}{\xmark} & OA (\%) & 81.57 & 56.14 & 60.68\\
        & & & & AA (\%) & 72.04 & 45.91 & 38.41\\
        & & & & Kappa (\%) & 70.24 & 42.39 & 50.05\\
        \hline
        \multirow{3}{*}{\cmark} & \multirow{3}{*}{\cmark} & \multirow{3}{*}{\cmark} & \multirow{3}{*}{\cmark} & OA (\%) & \bf 81.87 & \bf 56.56 & \bf 61.77\\
        & & & & AA (\%) & \bf 74.41 & \bf 46.13 & \bf 39.50\\
        & & & & Kappa (\%) & \bf 70.95 & \bf 42.67 & \bf 50.88\\
        \hline \hline
    \end{tabular}
    }
    \label{tab:ablation}
\end{table}

\section{Conclusion}
This paper proposes a multi-source collaborative domain generalization (MS-CDG) framework for cross-scene image classification in the remote sensing community. Benefiting from the homogeneity and heterogeneity characteristics of MS remote sensing data, it incorporates two aspects to enhance cross-scene generalization performance, namely data-aware adversarial augmentation and model-aware multi-level diversification. Specifically, the data-aware adversarial augmentation achieves multi-domain augmentation by effectively learning semantic changes and domain styles in the adversary neural network to expand high-quality MS samples and further achieve joint classification. The model-aware multi-level diversification explores the merits of both cross-domain and intra-domain class modeling by the pixel-to-prototype and KMM, which facilitates knowledge transfer of different domains and ensures better domain-invariant representation learning. Extensive experiments on three public MS remote sensing datasets demonstrate that our method is superior to state-of-the-art cross-scene image classification methods. In the future work, we would like to develop more advanced self-supervised strategies by introducing intrinsic physical characteristics of ground objects to improve the model's generalization ability.

\bibliographystyle{IEEEtran}
\bibliography{HZ_ref}

\end{document}